\documentclass{article}

\usepackage[utf8]{inputenc} 
\usepackage[T1]{fontenc}    
\usepackage[english]{babel}
\usepackage{url}            
\usepackage{booktabs}       
\usepackage{amsfonts}       
\usepackage{nicefrac}       
\usepackage{microtype}      
\usepackage{mathtools}
\usepackage{amsmath,amssymb}
\usepackage{amsthm}
\usepackage{thmtools,thm-restate}
\usepackage{mathrsfs}
\usepackage{float}
\usepackage{array}
\usepackage{graphicx}
\usepackage{caption}
\usepackage{enumitem}
\usepackage{subcaption}
\usepackage{array}
\usepackage[title]{appendix}
\usepackage{tikz}
\usetikzlibrary{positioning,angles,quotes}
\usepackage{xcolor}

\usepackage{trackchanges}
\addeditor{Adrien}

\usepackage{hyperref}

\newcommand{\montezuma}{\textsc{Montezuma's Revenge}}

\def \hN {\hat N}

\usepackage[accepted]{icml2019}

\icmltitlerunning{Benchmarking Bonus-Based Exploration Methods on the Arcade Learning Environment}

\begin{document}

\twocolumn[
\icmltitle{Benchmarking Bonus-Based Exploration Methods on the Arcade Learning Environment}




\begin{icmlauthorlist}
\icmlauthor{Adrien Ali Ta\"{i}ga}{mila,google}
\icmlauthor{William Fedus}{mila,google}
\icmlauthor{Marlos C. Machado}{google}
\icmlauthor{Aaron Courville}{mila,cifar}
\icmlauthor{Marc G. Bellemare}{google,cifar}
\end{icmlauthorlist}

\icmlaffiliation{mila}{MILA, Universit\'{e} de Montr\'{e}al}
\icmlaffiliation{google}{Google Research, Brain Team}
\icmlaffiliation{cifar}{CIFAR Fellow}

\icmlcorrespondingauthor{Adrien Ali Ta\"{i}ga}{adrien.ali.taiga@umontreal.ca}

\icmlkeywords{exploration, count based exploration, reinforcement learning}

\vskip 0.3in
]



\printAffiliationsAndNotice{}  

\begin{abstract}
This paper provides an empirical evaluation of recently developed exploration algorithms within the Arcade Learning Environment (ALE).
We study the use of different reward bonuses that incentives exploration in reinforcement learning. We do so by fixing the learning algorithm used and focusing only on the impact of the different exploration bonuses in the agent's performance. We use Rainbow, the state-of-the-art algorithm for value-based agents, and focus on some of the bonuses proposed in the last few years.
We consider the impact these algorithms have on performance within the popular game \montezuma{} which has gathered a lot of interest from the exploration community, across the the set of seven games identified by \citet{bellemare2016unifying} as challenging for exploration, and easier games where exploration is not an issue.
We find that, in our setting, recently developed bonuses do not provide significantly improved performance on \montezuma{}  or hard exploration games. We also find that existing bonus-based methods may negatively impact performance on games in which exploration is not an issue and may even perform worse than $\epsilon$-greedy exploration.

\end{abstract}

\section{Introduction}

Despite recent entreaties for better practices to yield reproducible research \citep{henderson2018deep,machado2018revisiting}, the literature on exploration in reinforcement learning still lacks a \emph{systematic} comparison between existing methods.
In the context of the the Arcade Learning Environment \citep[ALE;][]{bellemare2013arcade}, we observe comparisons of agents trained under different regimes: with or without reset, using varying number of training frames, with and without sticky actions \citep{machado2018revisiting} and evaluating only on a small subset of the available games.
This makes it nearly impossible to assess the field's progress towards efficient exploration.

Our goal here is to revisit some of the recent bonus based exploration methods using a common evaluation regime. We do so by
\begin{itemize}[leftmargin=0.9cm,topsep=0.5pt,itemsep=-1ex,partopsep=2ex,parsep=2ex]
	\item Comparing all methods on the same set of Atari 2600 games;
	\item Applying these bonuses on the same value-based agent architecture, Rainbow \citep{hessel2018rainbow};
	\item Fixing the number of samples each algorithm uses during training to 200 million game frames.
\end{itemize}
As an additional point of comparison, we also evaluate in the same setting NoisyNets \citep{fortunato18noisy}, part of the original Rainbow algorithm and $\epsilon$-greedy exploration.

We study three questions relevant to exploration in the ALE:
\begin{itemize}[leftmargin=0.9cm,topsep=0.5pt,itemsep=-1ex,partopsep=2ex,parsep=2ex]
	\item How well do different methods perform on \montezuma{}?
	\item Do these methods generalize to \citeauthor{bellemare2016unifying}'s set of ``hard exploration games'', when their hyperparameters are tuned only on \montezuma{}?
	\item Do they generalize to other Atari 2600 games?
\end{itemize}
We find that, despite frequent claims of state-of-the-art results in \montezuma{}, when the learning algorithm and sample complexity are kept fixed across the different methods, little to no performance gain can be observed over older methods.
Furthermore, our results suggest that performance on \montezuma{} is not indicative of performance on other hard exploration games. 
In fact, on 5 out of 6 hard exploration games performance of considered bonus-based methods is on-par with an $\epsilon$-greedy algorithm, and significantly lower than human-level performance.
Finally, we find that, while exploration bonuses improve performance on hard exploration games, they typically hurt performance on the easier Atari 2600 games.
Taken together, our results suggests that more research is needed to make bonus-based exploration robust and reliable, and serve as a reminder of the pitfalls of developing and evaluating methods primarily on a single domain.

\section{Related Work}
Exploration methods may encourage agents toward unexplored parts of the state space in different ways.
Count-based methods generalize previous work that was limited to tabular methods \citep{strehl2008analysis} to estimate counts in high dimension
\citep{bellemare2016unifying,count-based,tang2017exploration,choshen2018dora,machado18_SR}.
Prediction error has also been used as a novelty signal to compute an exploration bonus \citep{stadie2015incentivizing,pathak17curiositydriven,burda2018exploration}.
Another class of exploration methods apply the Thompson sampling heuristic to reinforcement learning \citep{osband2016deep,o2017uncertainty,touati2018randomized}.

\citet{burda2018large} benchmarks various exploration methods based on prediction error within a set of simulated environment including some Atari 2600 games.
However their study differs from ours as their setting ignore the environment reward and instead learns exclusively from the intrinsic reward signal.




\section{Exploration methods}
\label{sec:exploration_methods}
We focus on bonus-based methods that encourage exploration through a reward signal. 
At each time-step the agent is trained with the reward $r_t = e_t + \beta \cdot i_t$ where $e_t$ is the extrinsic reward provided by the environment, $i_t$ the intrinsic reward computed by agent and $\beta > 0$ a scaling parameter.
We now summarize different ways to compute the intrinsic reward $i$.

\subsection{Pseudo-counts}
Pseudo-counts \citep{bellemare2016unifying,count-based} were proposed as way to estimate counts in high dimension states spaces using a density model. The agent is then encouraged to visit states with a low visit count.
Let $\rho$ be a density model over the state space and $\rho_t (s)$ the density assigned to $s$ after being trained on a sequence of states $s_1, ..., s_t$.
We will write $\rho'_t (s)$ the density assigned to $s$ if $\rho$ were to be updated with $s$.
We require $\rho$ to be learning positive (i.e $\rho'_t (s) \geq \rho_t (s)$) and define the prediction gain as $\text{PG}_t (s) = \log \rho'_t (s) - \log \rho_t (s)$.
The pseudo-count $\hN_t (s_t) \approx \Big( e^{\text{PG}_t (s_t)} - 1 \Big)^{-1}$ can then be used to compute the intrinsic reward
\begin{equation}
\label{eq:psc}
    i^{\text{PSC}} (s_t) \coloneqq ( \hN_t (s_t) )^{-1/2}.
\end{equation}
CTS \citep{bellemare2014skip} and PixelCNN \citep{oord2016pixel} have been both used as density models. 
We will disambiguate these agent by the name of their density model.

\subsection{Intrinsic Curiosity Module}
Intrinsic Curiosity Module \citep[ICM,][]{pathak17curiositydriven} promotes exploration via curiosity. \citeauthor{pathak17curiositydriven} formulates curiosity as the agent's ability to predict the consequence of its own actions in a learned feature space. ICM includes a learned embedding, a forward and an inverse model.
The embedding is trained through the inverse model, which in turn, has to predict the agent's action between two states $s_t$ and $s_{t+1}$ using their embedding $\phi (s_t)$ and $\phi (s_{t+1})$.
Given a transition $(s_t, a_t, s_{t+1})$ the intrinsic reward is then given by the error of the forward model in the embedding space between $\phi (s_{t+1})$ and the predicted estimate $\hat{\phi} (s_{t+1})$
\begin{equation}
\label{eq:icm}
    i^{\text{ICM}} (s_t) = \| \hat{\phi} (s_{t+1}) - \phi (s_{t+1}) \|_2^2 .
\end{equation}

\subsection{Random Network Distillation}
Random Network Distillation \citep[RND,][]{burda2018exploration} derives a bonus from the prediction error of a random network.
The intuition is that the prediction error will be low on states that are similar to those previously visited and high on newly visited states?.
A neural network $\hat f$ with parameters $\theta$ is trained to predict the output of a fixed randomly initialized neural network $f$:
\begin{equation}
\label{eq:rnd}
    i^{\text{RND}} (s_t) = \| \hat{f} (s_t; \theta) - f(s_t) \|_2^2
\end{equation}

\subsection{NoisyNets}
Though is does not generate an exploration bonus, we also evaluate NoisyNets \citep{fortunato18noisy} as it was chosen as the exploration strategy of the original Rainbow implementation \citep{hessel2018rainbow}. NoisyNets add noise in parameter space and propose to replace standard fully-connected layers $y = Ws + b$ by a noisy version that combines a deterministic and a noisy stream:
\begin{equation}
    y = (W + W_{noisy} \odot \epsilon^W) s + (b + b_{noisy} \odot \epsilon^b),
\end{equation}
where $\epsilon^W$ and $\epsilon^b$ are random variable and $\odot$ denotes elementwise multiplication.

\section{Evaluation protocol}

We evaluate two key properties of exploration methods in the ALE:
\begin{itemize}
    \item \textbf{Sample efficiency}: obtaining a decent policy quickly.
    \item \textbf{Robustness}: performing well across different games of the ALE with the same set of hyperparameters.
\end{itemize}
Sample efficiency is a key objective for exploration methods, yet, because published agents are often trained under different regimes it is often not possible to directly compare their performance.
They often employ different reinforcement learning algorithms, varying quantity of training frames or inconsistent hyperparameter tuning.
As a remedy, we fix our training protocol and train bonus-based methods with a common agent, the Rainbow implementation provided by the Dopamine framework \citep{castro2018dopamine} which includes Rainbow's three most important component: $n$-step updates \citep{mnih2016asynchronous}, prioritized experience replay \citep{schaul2015prioritized} and distributional reinforcement
learning \citep{bellemare2017distributional}.
To avoid introducing bias in favor of a particular method we also kept the original hyperparameters fixed.
Our agents are trained for 200 million frames following \citeauthor{mnih2015human}'s original setting.
Nevertheless we also acknowledge the emerging trend of training agents an order of magnitude longer in order to produce a high-scoring policy, irrespective of the sample cost \citep{Espeholt2018IMPALASD,burda2018exploration,kapturowski2018recurrent}.

The ALE was designed with the assumption that few games would be used for training and the remaining ones for evaluation.
Nonetheless it has become common to do hyperparameter tuning on \textsc{Montezuma's Revenge} and only evaluate on other ALE's hard exploration games with sparse rewards: \textsc{Freeway}, \textsc{Gravitar}, \textsc{Solaris}, \textsc{Venture}, \textsc{Private Eye}.
While this may be due to limited computational resources doing so however may come to a price on easier exploration problems as we will see later on.
For this reason we chose to also evaluate performance on the original Atari training set\footnote{\textsc{Freeway}, \textsc{Asterix}, \textsc{Beam Rider}, \textsc{Seaquest}, \textsc{Space Invaders}}. 
Except for \textsc{Freeway} these are all considered easy exploration problems \citep{bellemare2016unifying}.

\begin{figure}
\centering
\includegraphics[width=\linewidth]{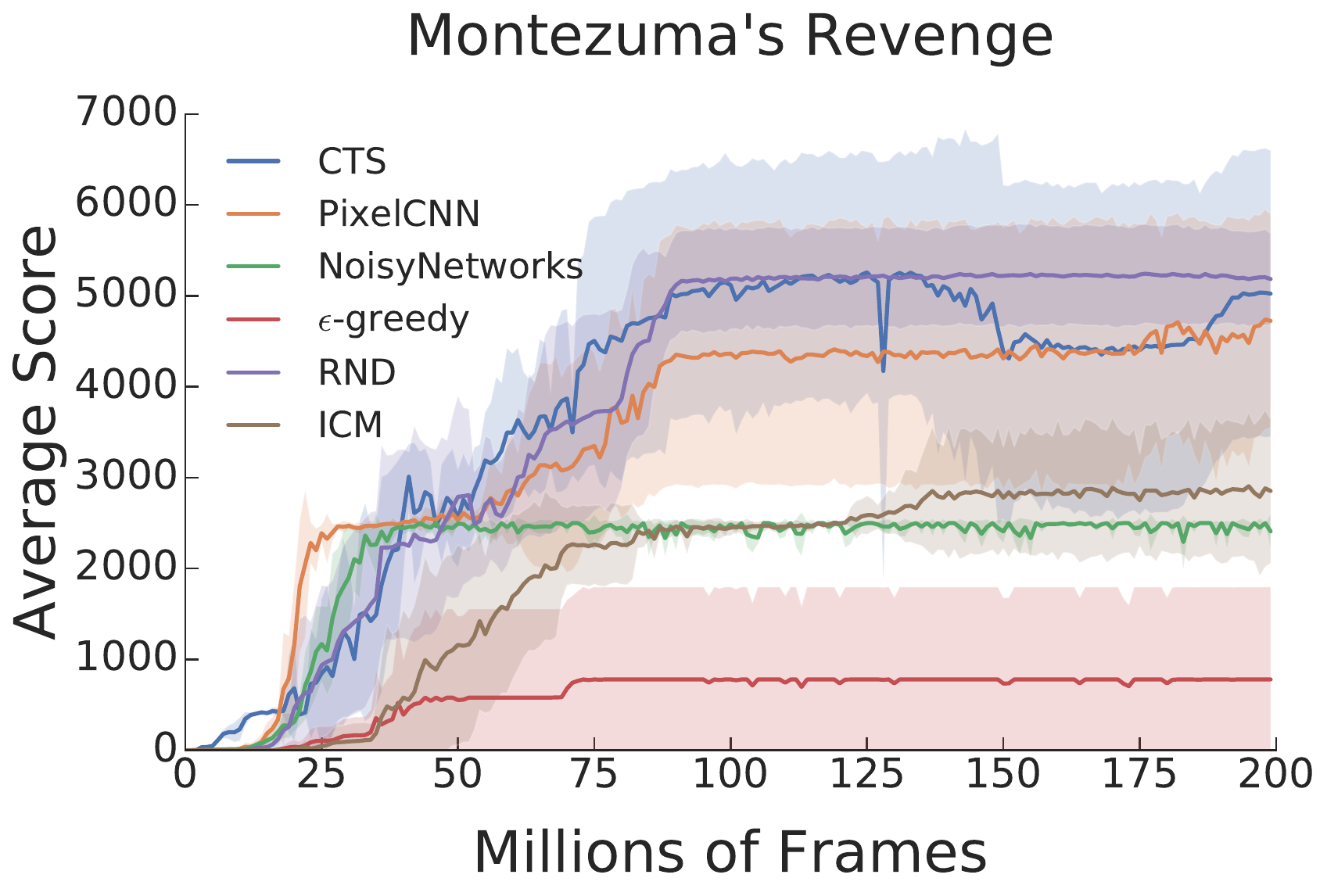}
\caption{A comparison of different exploration methods on \montezuma{}.} 
\label{fig:montezuma}
\end{figure}

\section{Empirical Results}
In this section we present an experimental study of exploration methods using the protocol described previously.

\subsection{\textsc{Montezuma's revenge}}
We begin by establishing a benchmark of bonus-based methods on \montezuma{} when each method is tuned on the same game.
Details regarding implementation and hyperparameter tuning may be found in Appendix \ref{app:hparam_tuning}.
Figure~\ref{fig:montezuma} shows training curves (averaged over 5 random seeds) for Rainbow augmented with different exploration bonuses.

As anticipated, $\epsilon$-greedy exploration performs poorly. 
Other strategies are able to consistently reach 2500 points and often make further progress.
We find pseudo-count with CTS matches recent bonuses and reaches a score of 5000 points within 200 millions frames. 
Of note, the performance we report for each method improves on the performance originally reported by the authors. This is mostly due to the fact these methods are based on weaker Deep Q-Network \citep{mnih2015human} variants. This emphasize again the importance of the agent architecture to evaluate exploration methods.

Regarding RND performance, we note that our implementation only uses Eq. \eqref{eq:rnd} bonus and does not appeal to other techniques presented in the same paper that were shown to be critical to the final performance of the algorithm. Though, we might expect that such techniques would also benefit other bonus based methods and leave it to future work. 

\begin{figure*}
\centering
\captionsetup{justification=centering}
\begin{subfigure}{0.25\textwidth}
\centering
  \includegraphics[width=\linewidth]{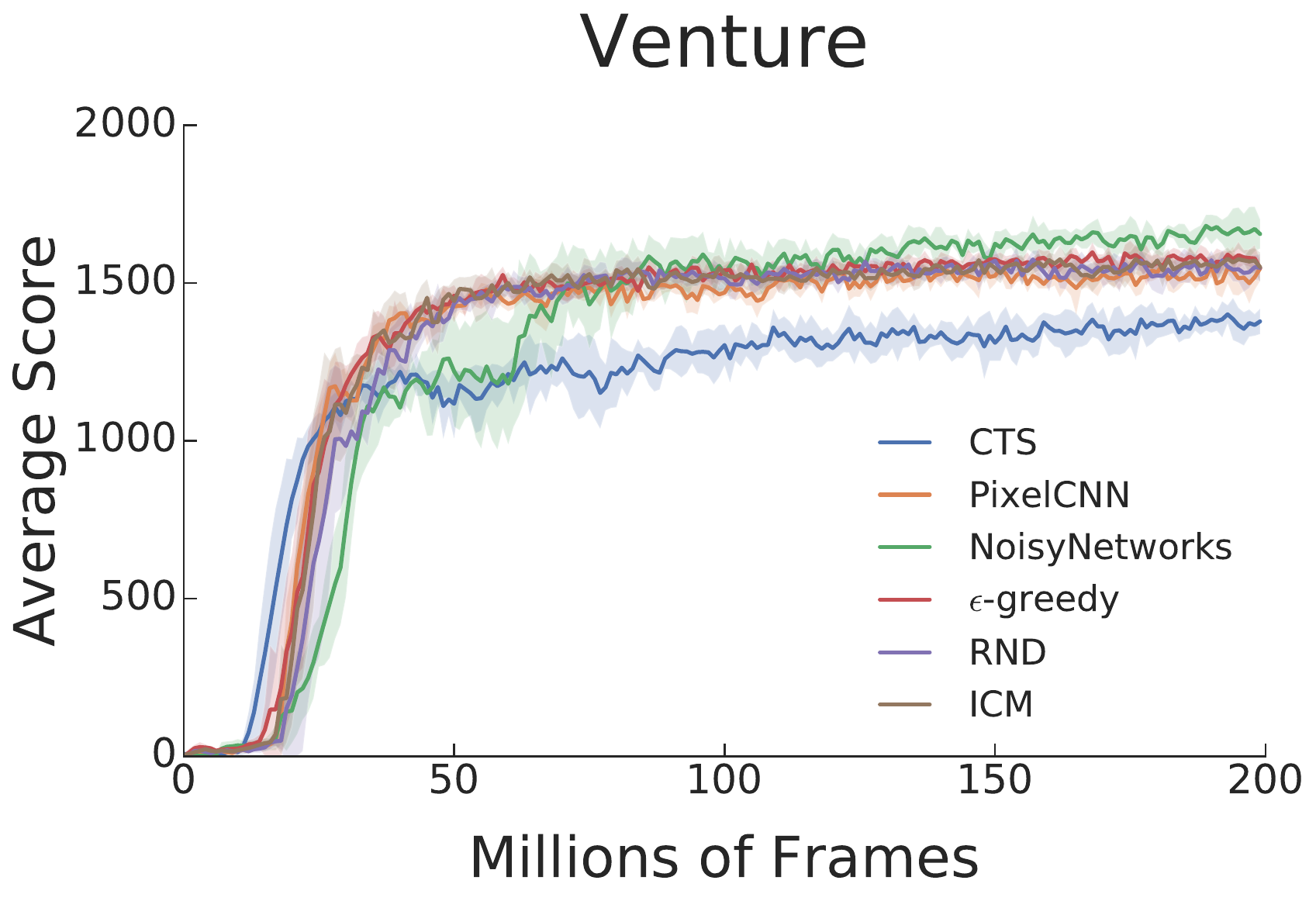}
  \label{fig:ven}
\end{subfigure}%
\begin{subfigure}{0.25\textwidth}
\centering
  \includegraphics[width=\linewidth]{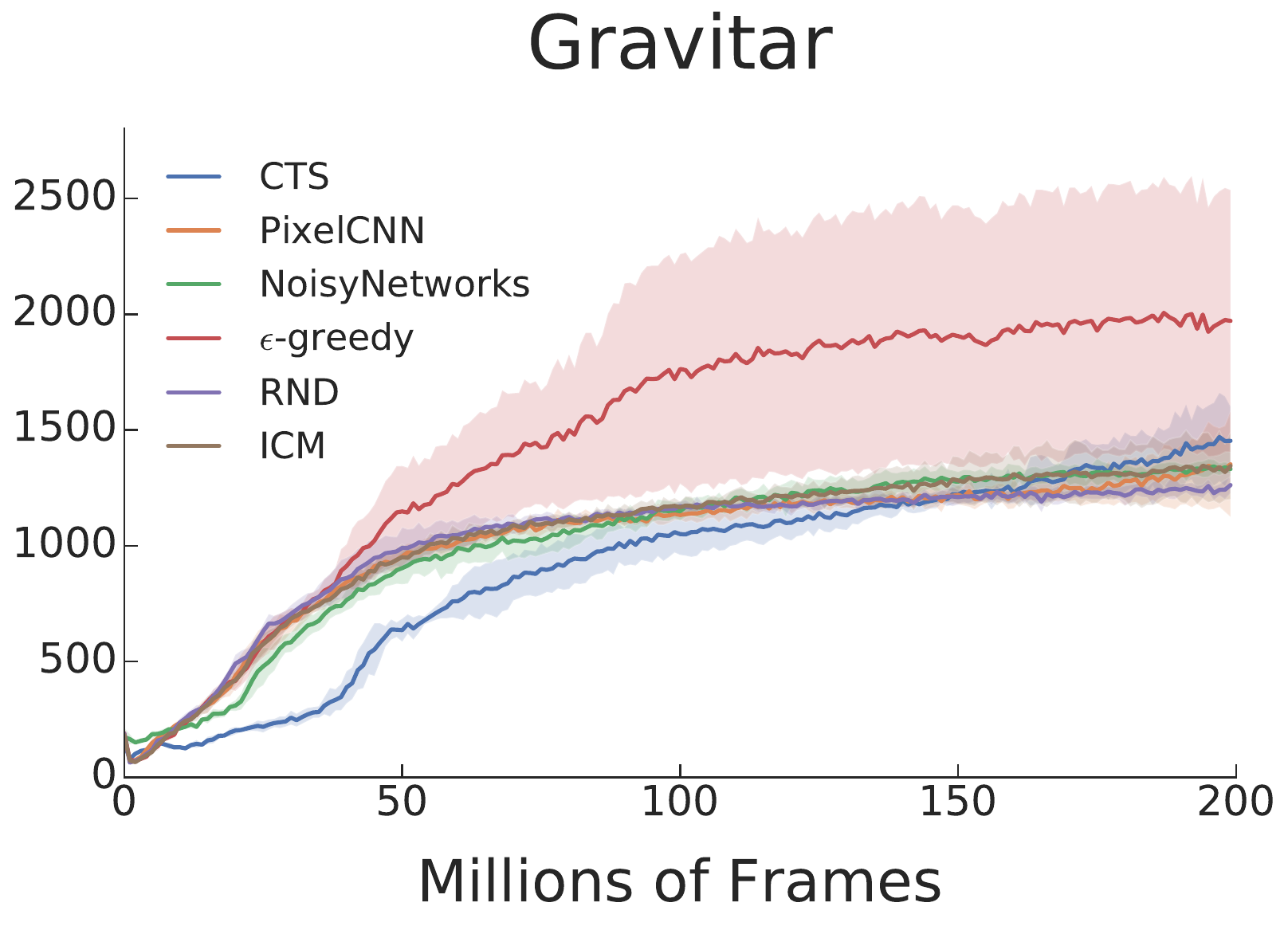}
  \label{fig:grav}
\end{subfigure}%
\begin{subfigure}{0.25\textwidth}
    \centering
  \includegraphics[width=\linewidth]{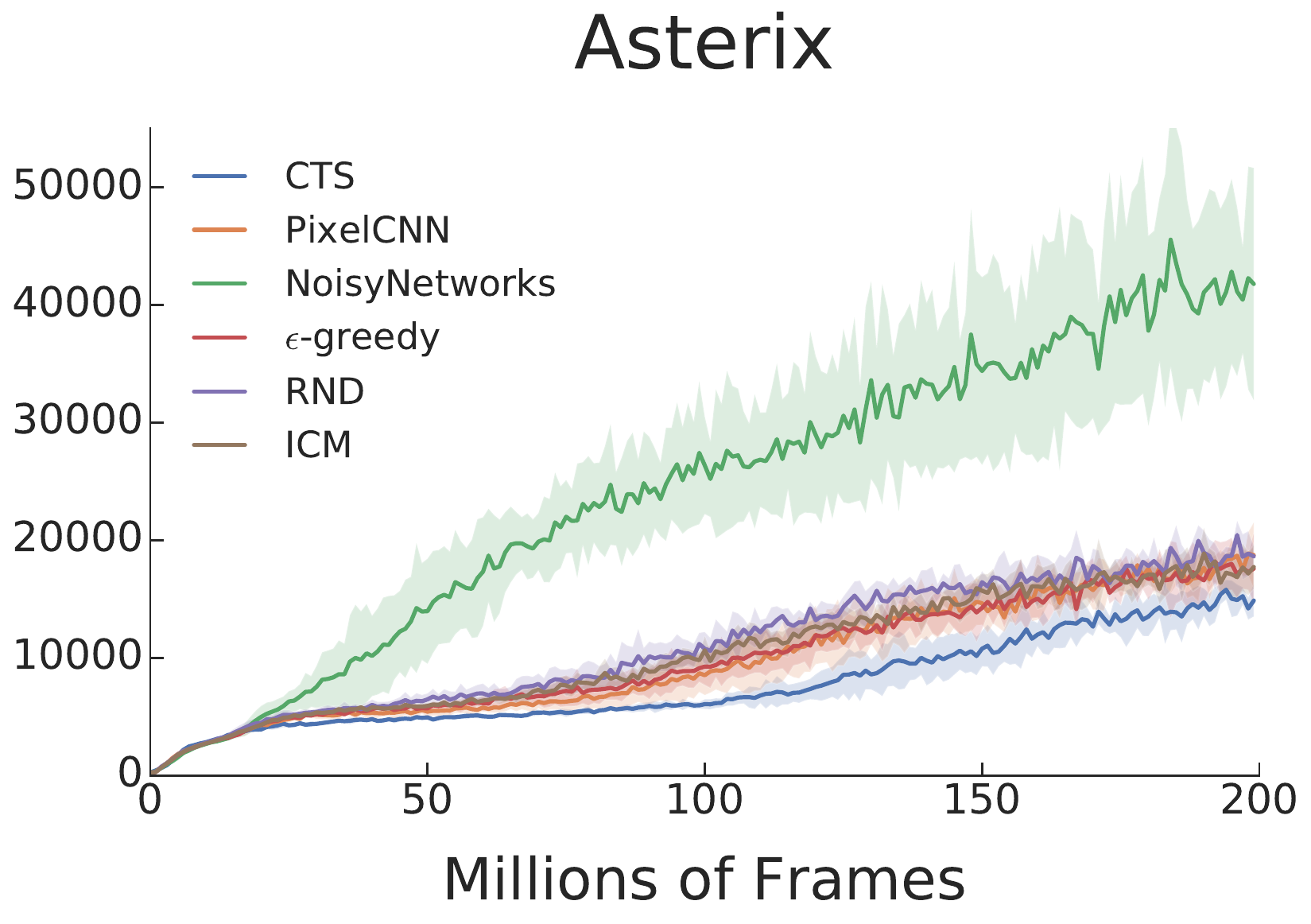}
  \label{fig:ast}
\end{subfigure}%
\begin{subfigure}{0.25\textwidth}
\centering
  \includegraphics[width=\linewidth]{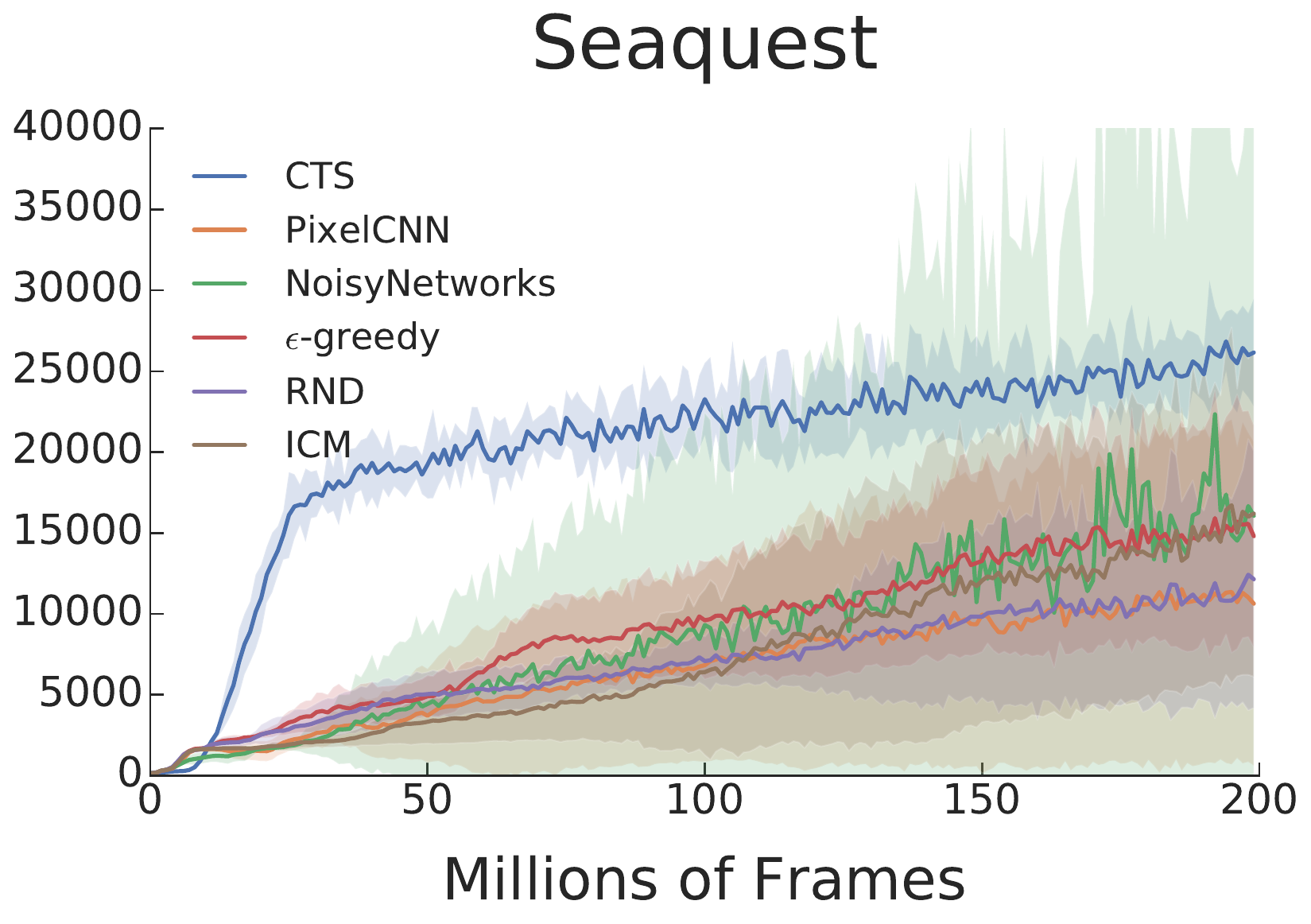}
  \label{fig:sea}
\end{subfigure}%
\caption{Evaluation of different bonus-based exploration methods on the ALE. \textsc{Venture} and \textsc{Gravitar} are hard exploration games whereas \textsc{Asterix} and \textsc{Seaquest} are easy ones.
}
\label{fig:evaluation_atari}
\end{figure*}

\subsection{Hard exploration games}
We now turn our attention to the set of games categorized as \emph{hard exploration games} by \citet{bellemare2016unifying} that is often used as an evaluation set for exploration methods.
Training curves for few games are shown in Figure~\ref{fig:evaluation_atari}, the remaining ones are in Appendix \ref{sec:curves}.
We find that performance of each method on \montezuma{} does not correlate with performance on other hard exploration problems and the gap between different methods is not as large as it was on \montezuma{}.
Surprisingly, in our setting, there is also no visible difference between $\epsilon$-greedy exploration and more sophisticated exploration strategies.
$\epsilon$-greedy exploration remains competitive and even outperforms other methods by a significant margin on \textsc{Gravitar}.
Similar results have been reported previously \citep{machado18_SR,burda2018exploration}.
These games were originally classified as hard exploration problems because DQN with $\epsilon$-greedy exploration was unable to reach a high scoring policy; however, these conclusions must be revisited with stronger base agents.
Progress in these games may be due to better credit assignment methods and not to the underlying exploration bonus.

\subsection{ALE training set}
While the benefit of exploration bonuses has been shown on a few games they can also have a negative impact by skewing the reward landscape.
To get a more complete picture, we also evaluated our agents on the original Atari training set which includes many easy exploration games.
Figure~\ref{fig:evaluation_atari} shows training curves for \textsc{Asterix} and \textsc{Seaquest}, the remaining games can be found in Appendix \ref{sec:curves}.
In this setting we noticed a reversed trend than the one observed on \montezuma{}. 
The pseudo-count method ends up performing worse on every game except \textsc{Seaquest}.
RND and ICM are able to consistently match the level of $\epsilon$-greedy exploration, but not exceed it.
The earlier benefits conferred by pseudo-counts result in a considerable detriment when the exploration problem is not difficult.
Finally, since NoisyNets optimizes the true environment reward, and not a proxy reward, it consistently matches $\epsilon$-greedy and occasionally outperforms.
Overall we found that bonus-based methods are generally detrimental in the context of easy exploration problems.
Despite its limited performance on \montezuma{} NoisyNets gave the most consistent results across our evaluation despite its limited performance on \montezuma{}.

\section{Conclusion}
Many exploration methods in reinforcement learning are introduced with confounding factors -- longer training duration, different model architecture and new hyper parameters.
This obscures the underlying signal of the exploration method.
Therefore, following a growing trend in the reinforcement learning community, we advocate for better practices on empirical evaluation for exploration to fairly assess the contribution of newly proposed methods.
In a standardized training environment and context, we found that $\epsilon$-greedy exploration can often compete with more elaborate methods on the ALE.
This shows that more work is still needed to address the exploration problem in complex environments.

\section{Acknowledgements}
The authors would like to thank Hugo Larochelle, Benjamin Eysenbach, Danijar Hafner and Ahmed Touati for insightful discussions as well as Sylvain Gelly for careful reading and comments on an earlier draft of this paper.

\bibliography{paper}
\bibliographystyle{icml2019}

\newpage
\begin{appendices}
\onecolumn

\section{Additional figures}
\label{sec:curves}

\begin{figure*}[!ht]
\centering
\captionsetup{justification=centering}
\begin{subfigure}{0.33\textwidth}
\centering
  \includegraphics[width=\linewidth]{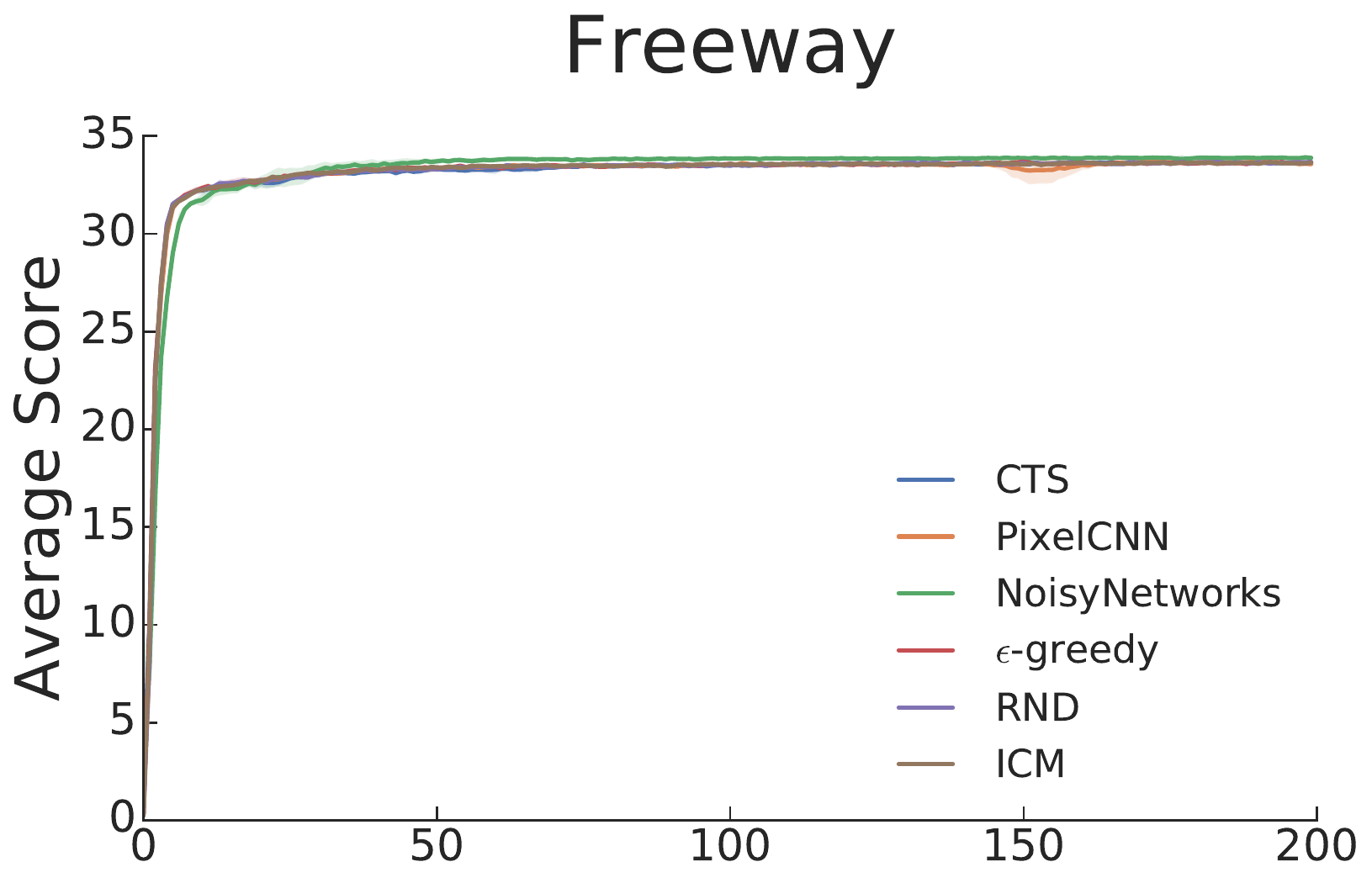}
  \label{fig:freeway}
\end{subfigure}%
\begin{subfigure}{0.33\textwidth}
\centering
  \includegraphics[width=\linewidth]{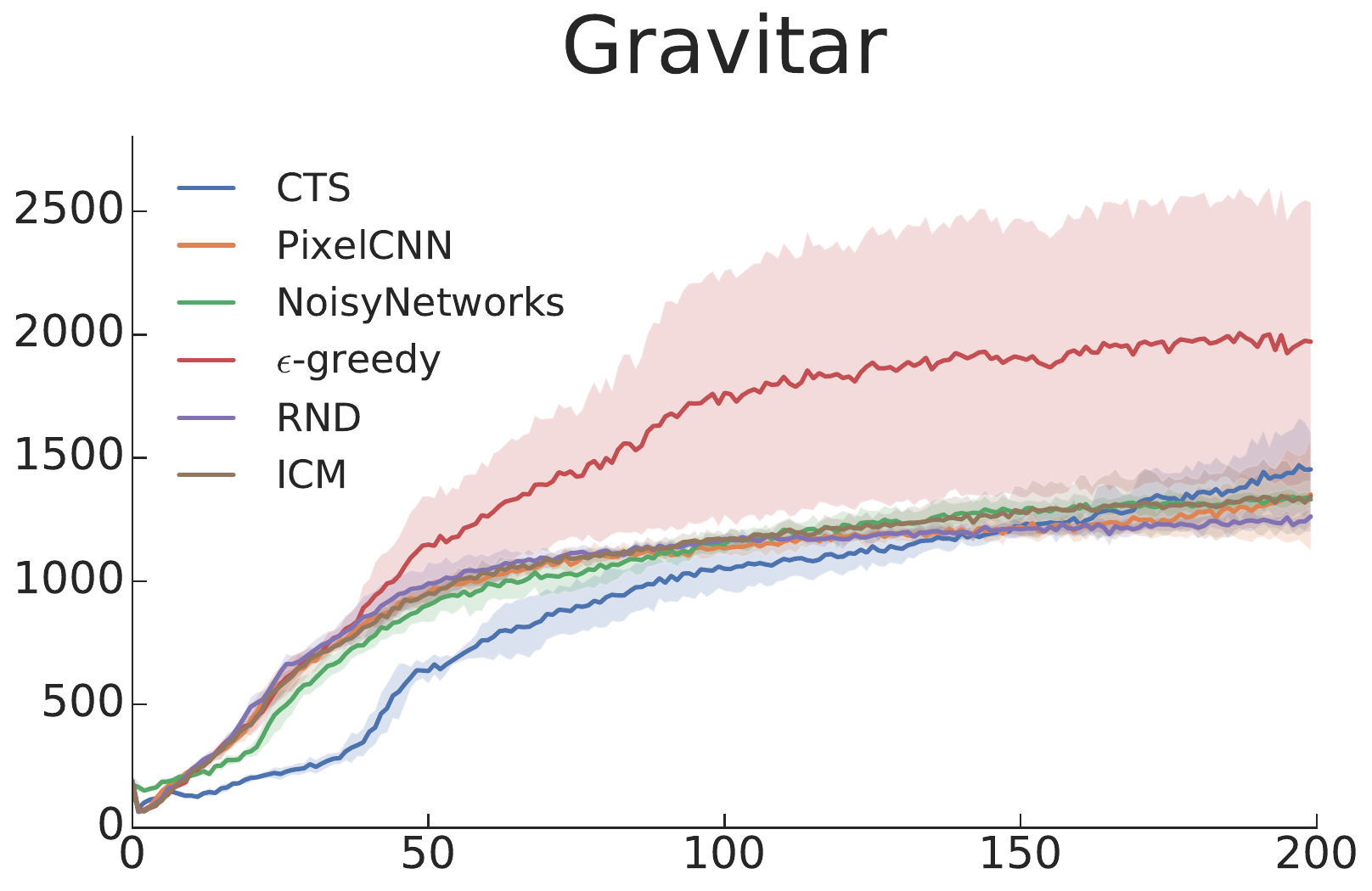}
  \label{fig:gravitar}
\end{subfigure}%
\begin{subfigure}{0.33\textwidth}
    \centering
  \includegraphics[width=\linewidth]{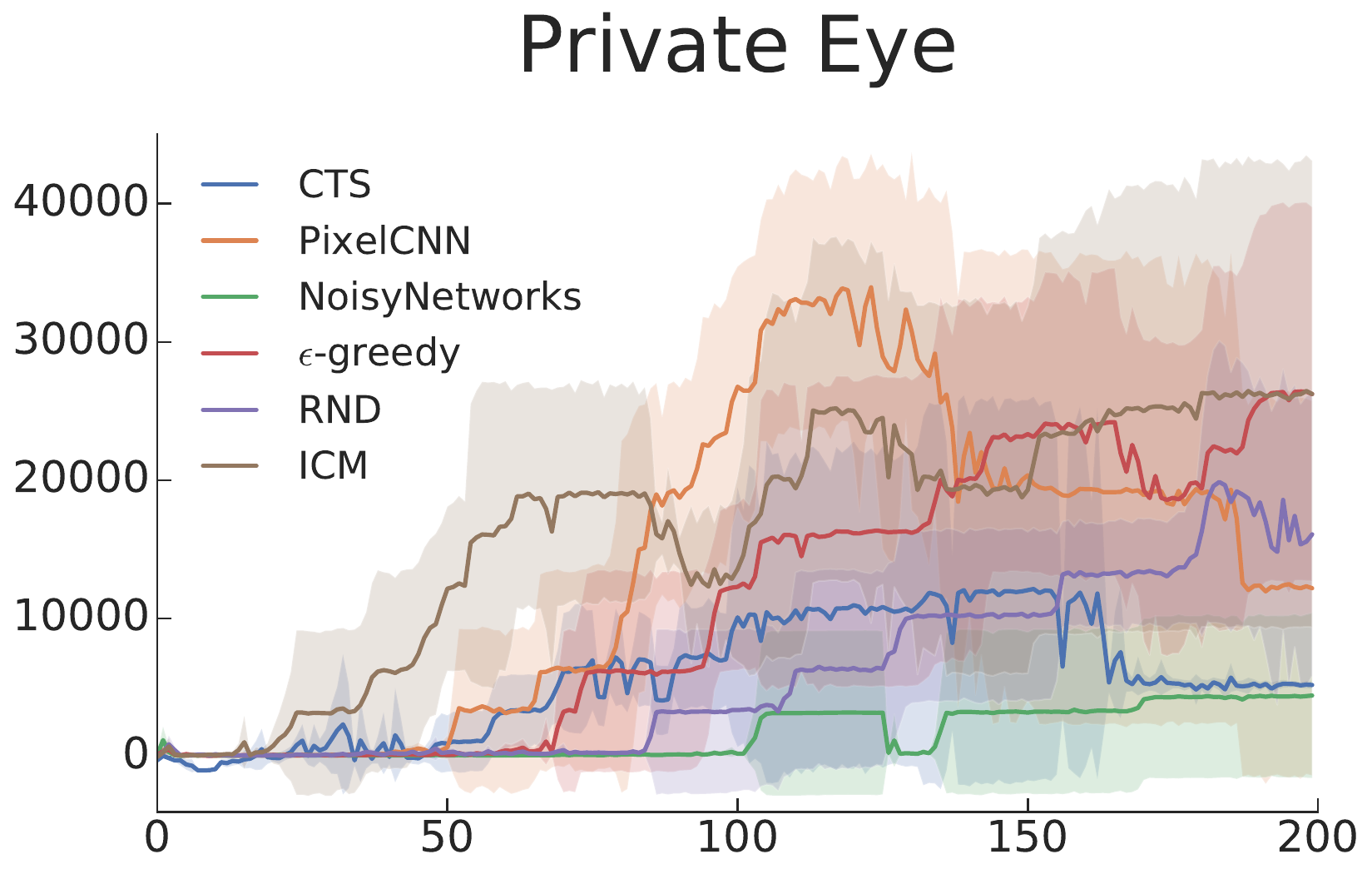}
  \label{fig:privateeye}
\end{subfigure}%
\vspace{4mm}
\begin{subfigure}{0.33\textwidth}
\centering
  \includegraphics[width=\linewidth]{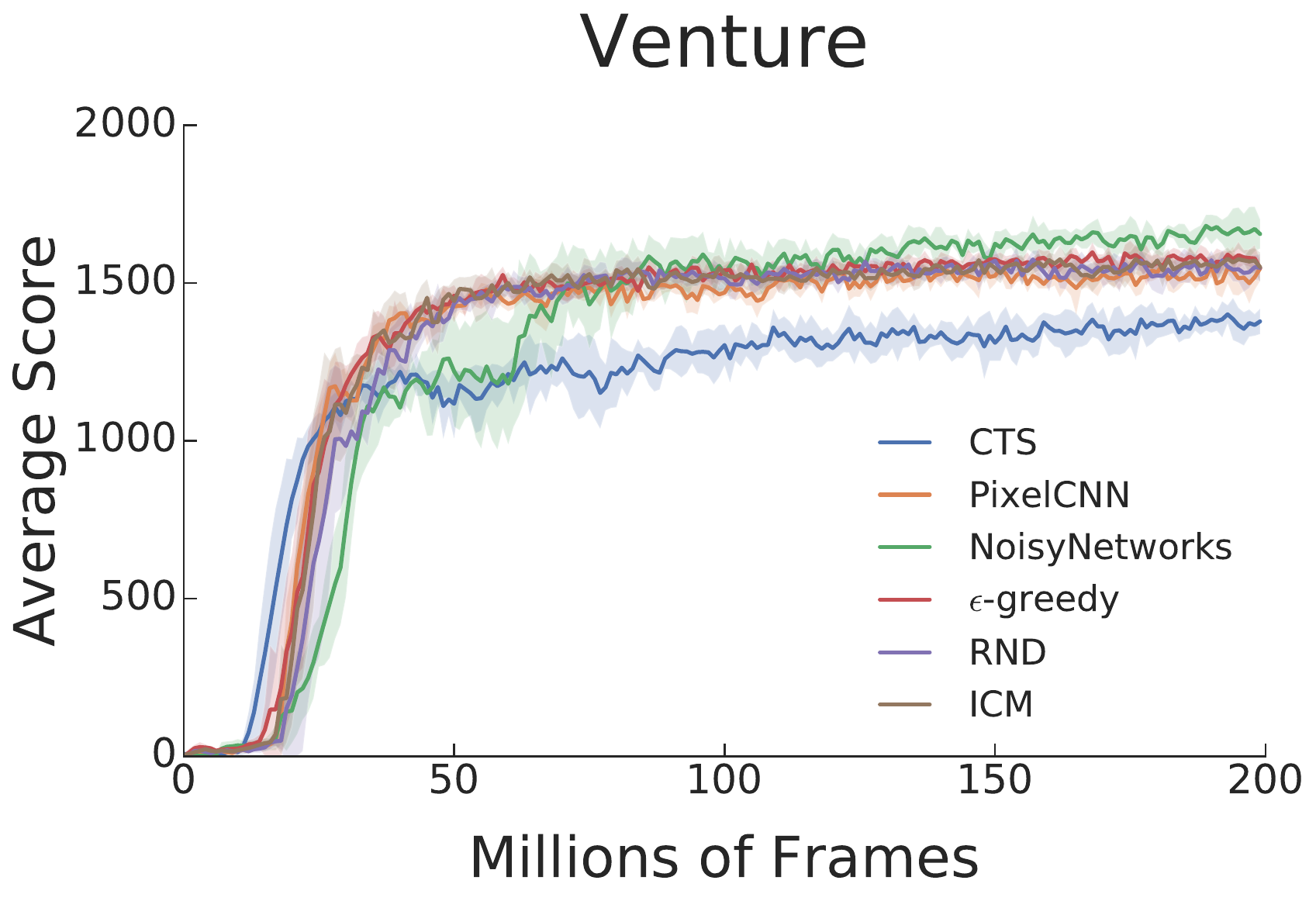}
  \label{fig:venture}
\end{subfigure}%
\begin{subfigure}{0.33\textwidth}
\centering
  \includegraphics[width=\linewidth]{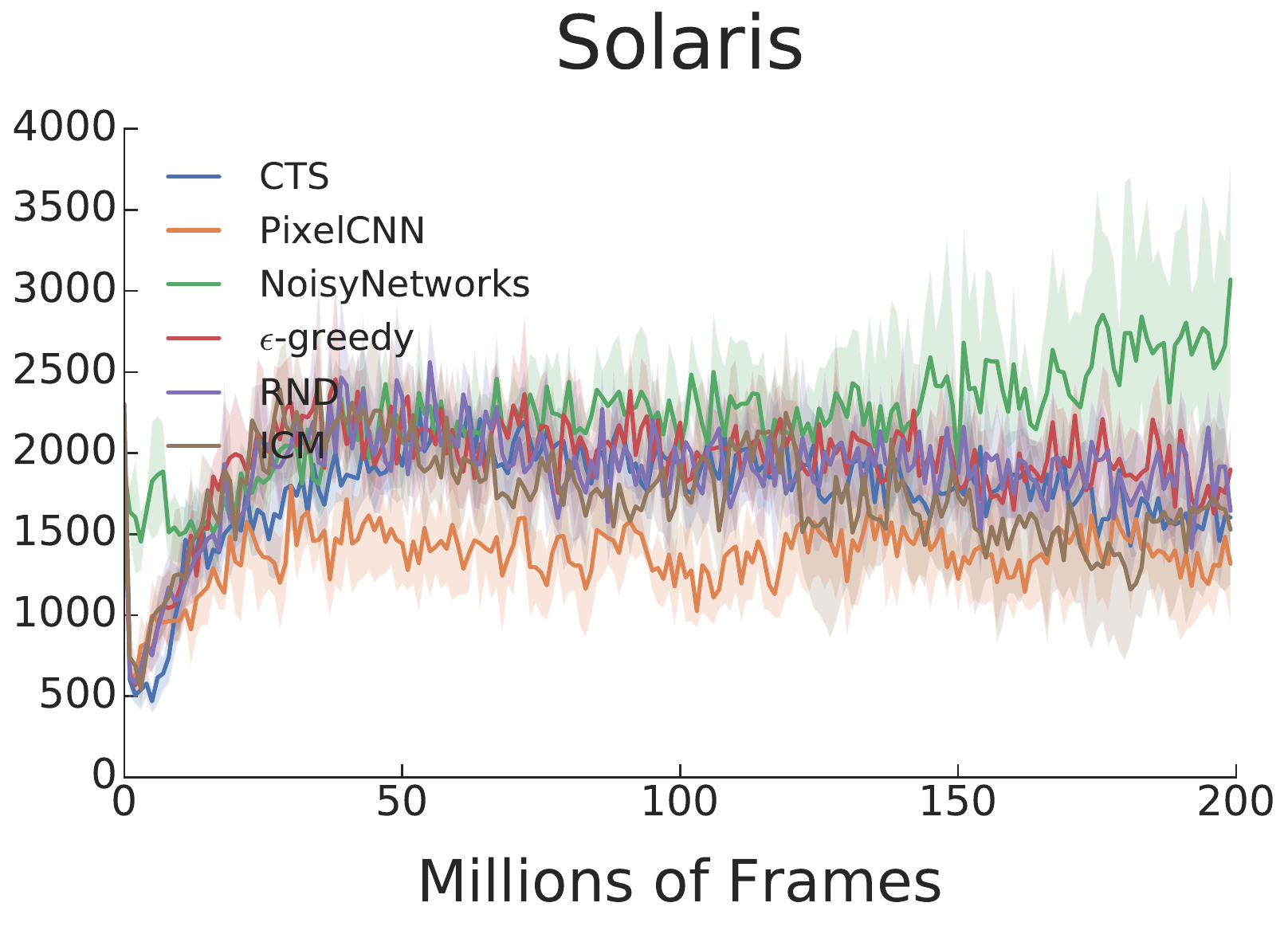}
  \label{fig:solaris}
\end{subfigure}%
\begin{subfigure}{0.33\textwidth}
\centering
  \includegraphics[width=\linewidth]{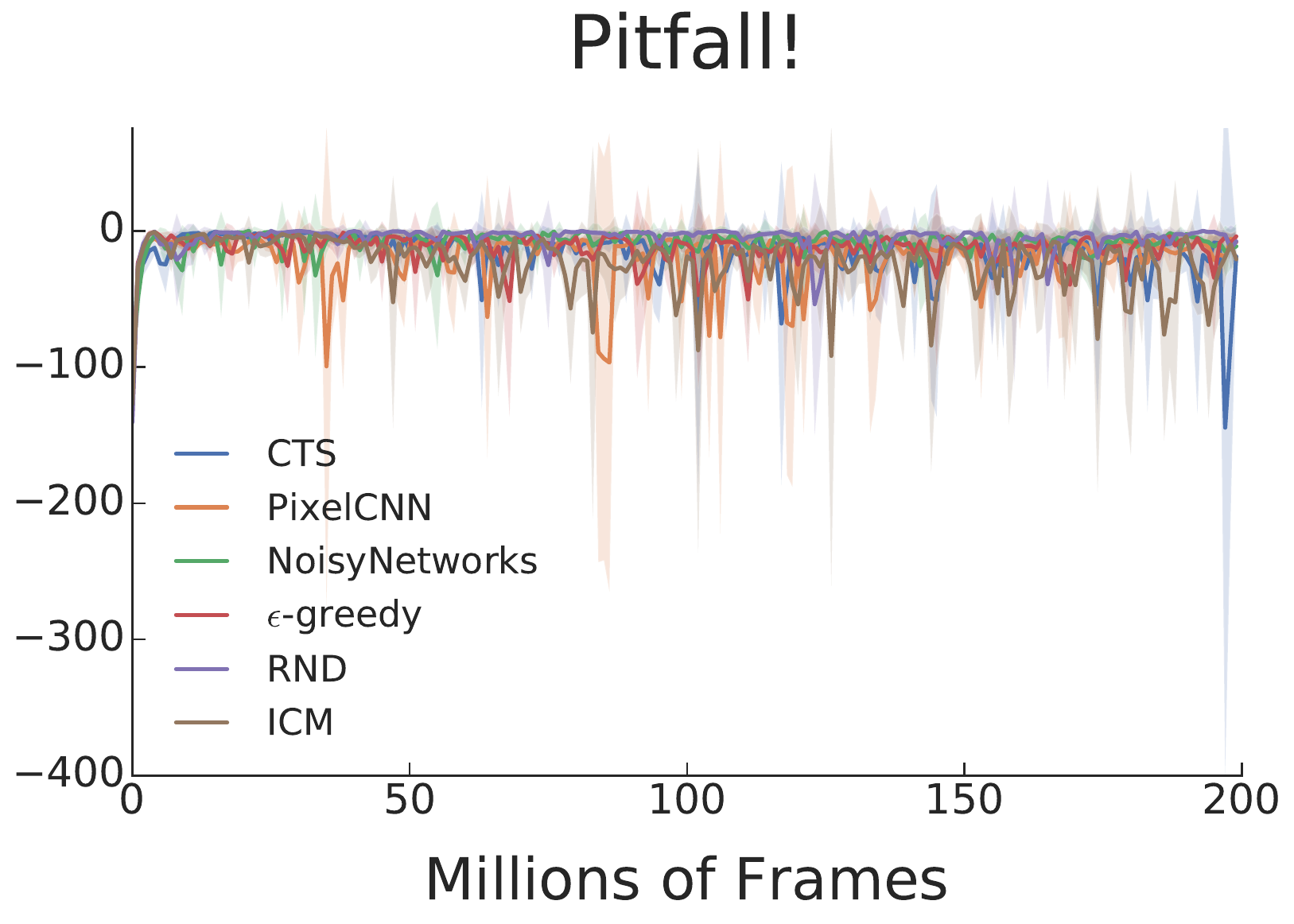}
  \label{fig:pitfall}
\end{subfigure}%
\caption{Results of different bonus-based exploration methods on hard exploration games. The relative ranking of methods differs from the one observed on \montezuma{}. We find that $\epsilon$-greedy also performs competitively.  This suggests that previous claims of progress in these games has been driven by more advanced reinforcement learning algorithms, not necessarily better exploration strategies.} 
\label{fig:hard}
\end{figure*}

\begin{figure*}[!ht]
\centering
\captionsetup{justification=centering}
\begin{subfigure}{0.33\textwidth}
    \centering
  \includegraphics[width=\linewidth]{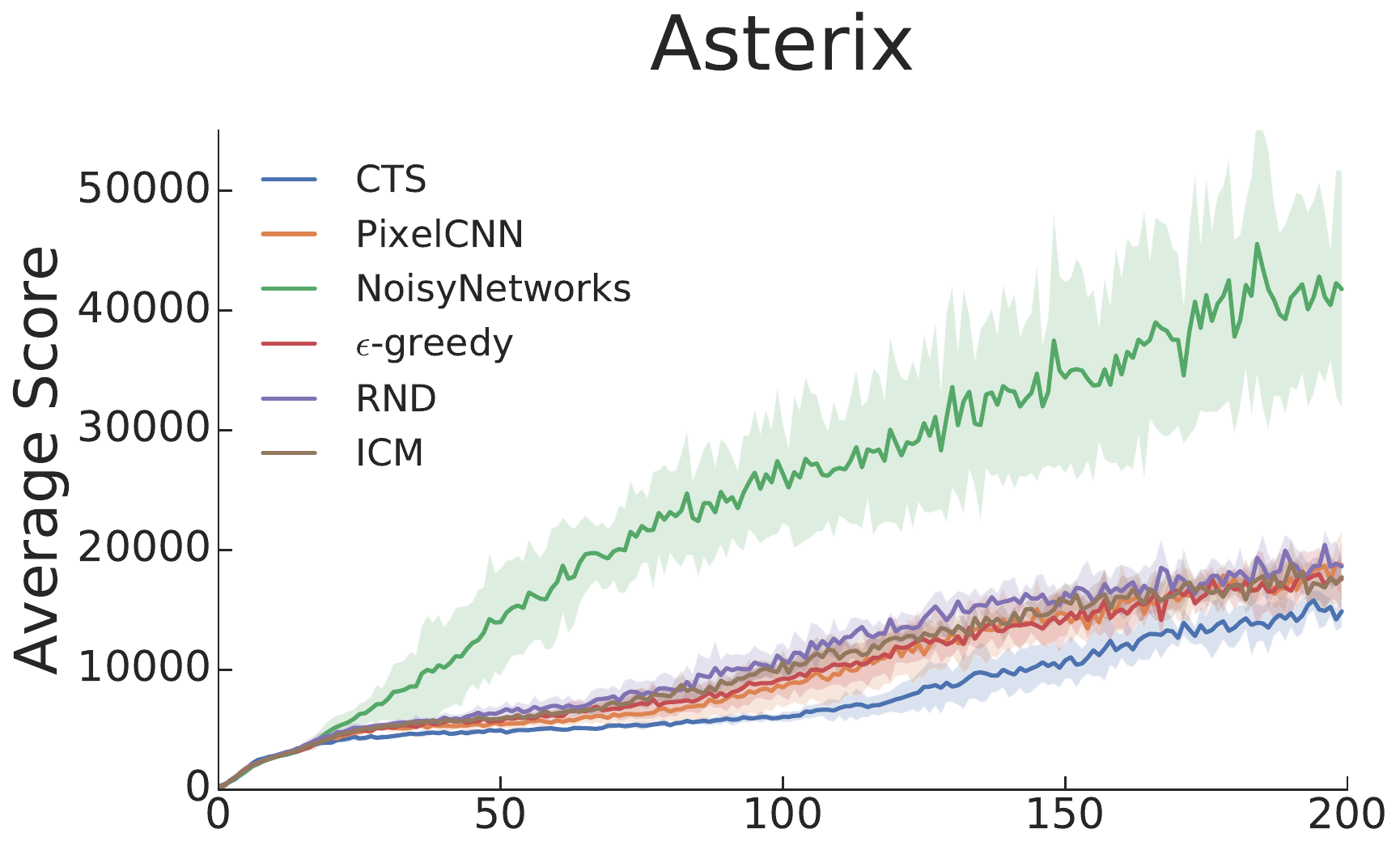}
  \label{fig:asterix}
\end{subfigure}%
\begin{subfigure}{0.33\textwidth}
\centering
  \includegraphics[width=\linewidth]{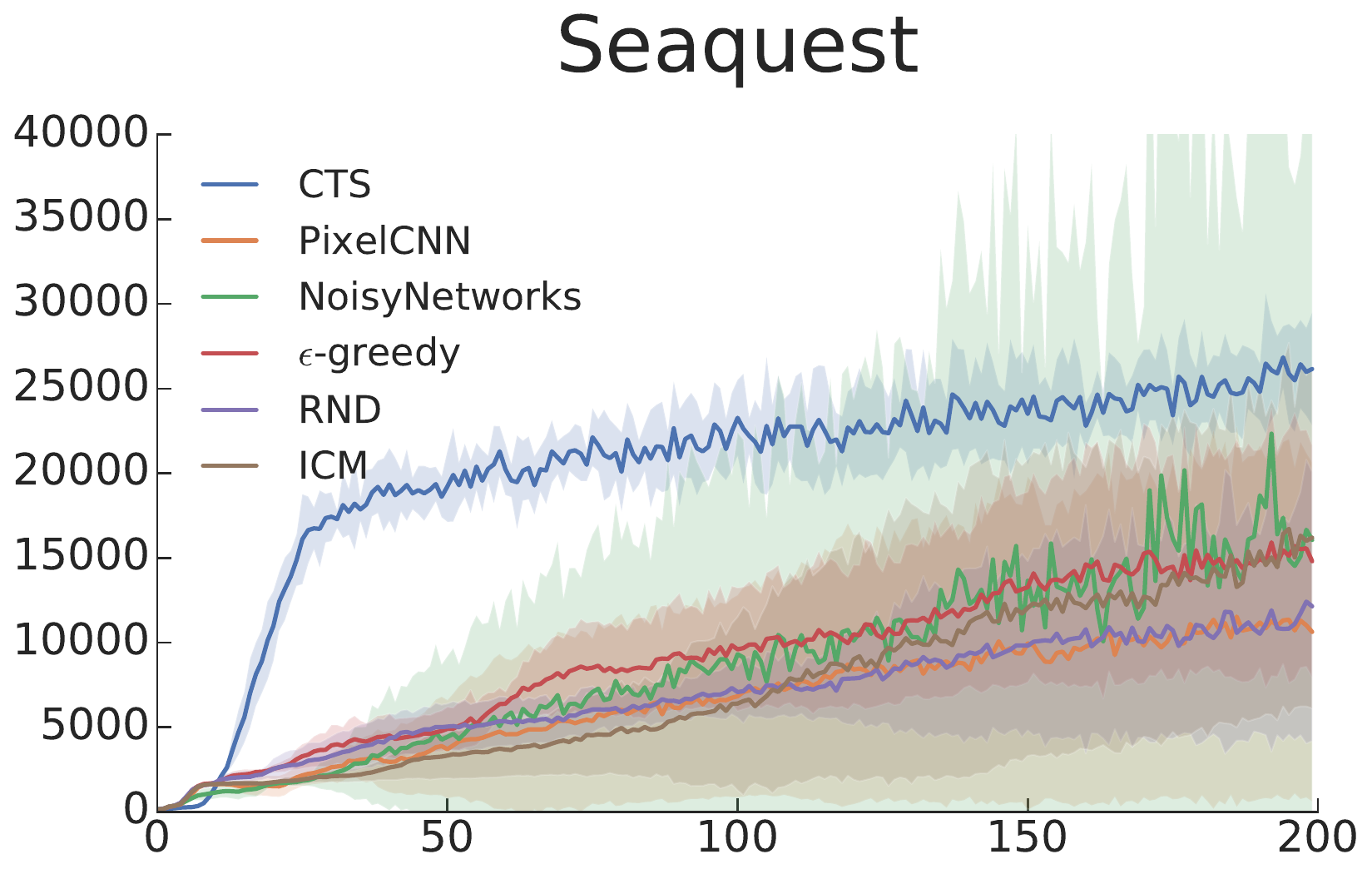}
  \label{fig:seaquest}
\end{subfigure}%
\begin{subfigure}{0.33\textwidth}
\centering
  \includegraphics[width=\linewidth]{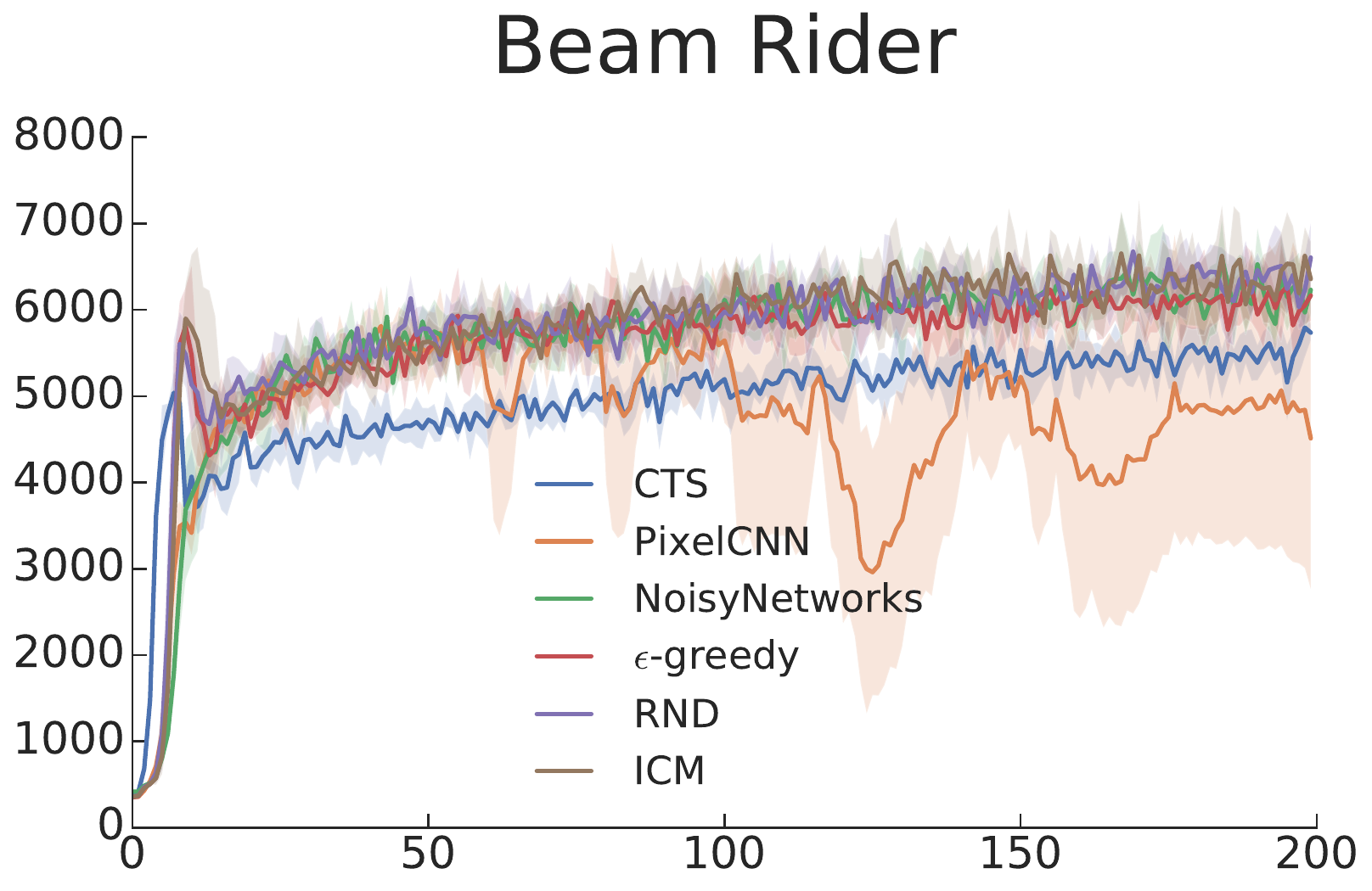}
  \label{fig:beamrider}
\end{subfigure}%
\vspace{4mm}
\begin{subfigure}{0.33\textwidth}
\centering
  \includegraphics[width=\linewidth]{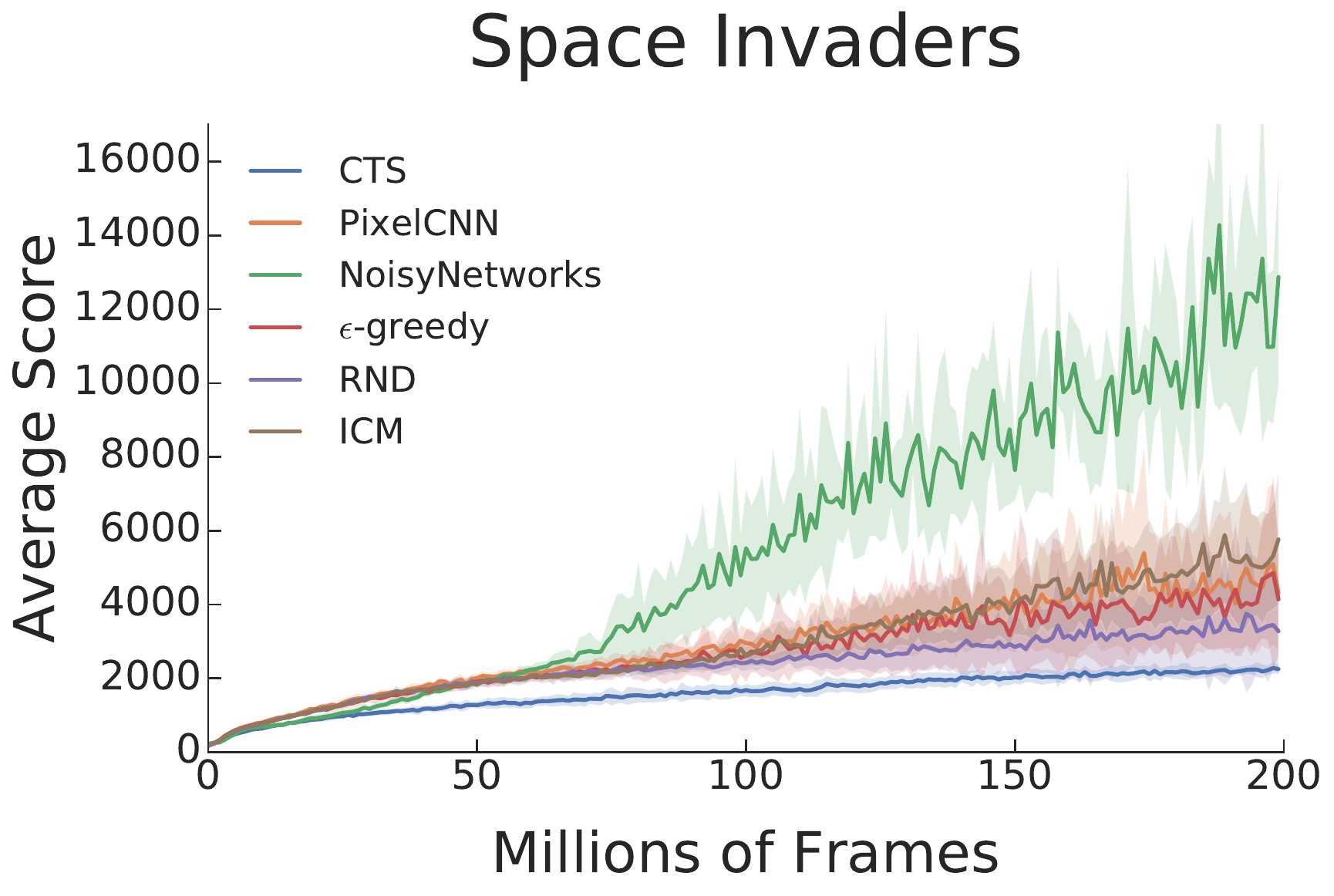}
  \label{fig:spaceinvaders}
\end{subfigure}%
\begin{subfigure}{0.33\textwidth}
\centering
  \includegraphics[width=\linewidth]{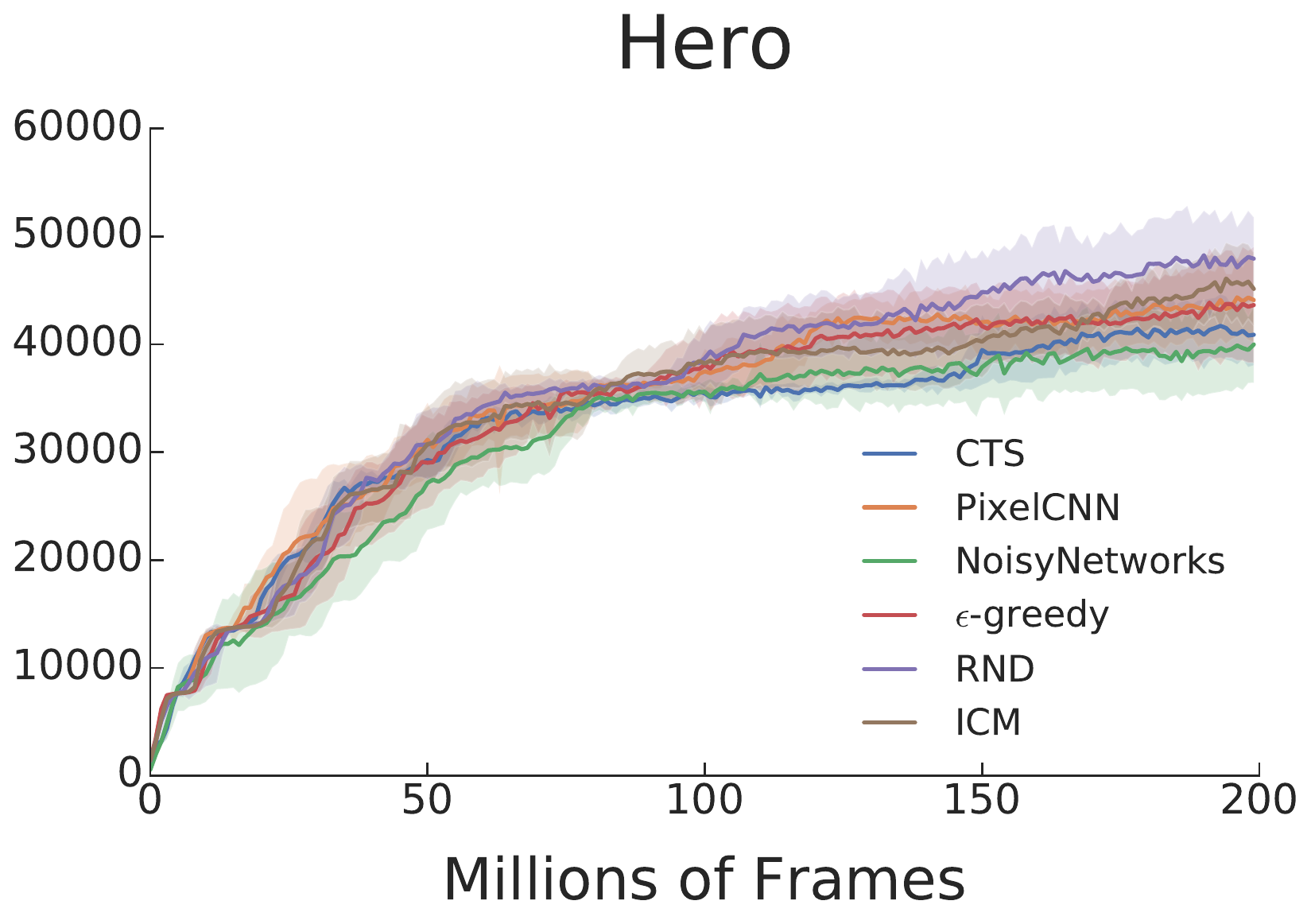}
  \label{fig:hero}
\end{subfigure}%
\begin{subfigure}{0.33\textwidth}
\centering
  \includegraphics[width=\linewidth]{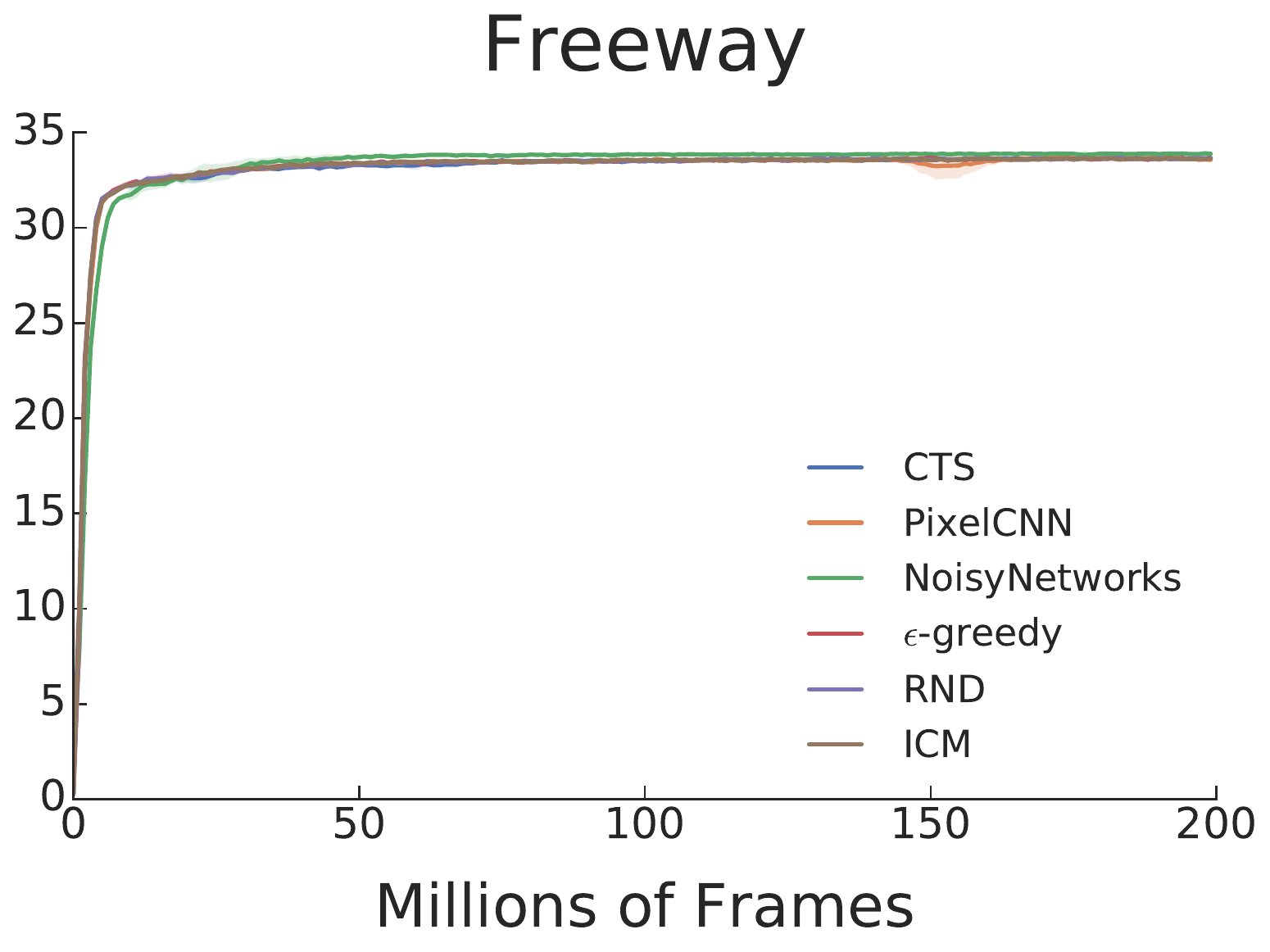}
  \label{fig:freeway2}
\end{subfigure}%
\caption{Evaluation of different bonus-based exploration methods on the Atari training set, except for \textsc{Freeway} all these games were classified as easy exploration problems. Rainbow with $\epsilon$-greedy exploration performs as well as other more complex exploration method.
}
\label{fig:training}
\end{figure*}

\begin{figure*}[!ht]
\centering
\captionsetup{justification=centering}
\begin{subfigure}{0.33\textwidth}
    \centering
  \includegraphics[width=\linewidth]{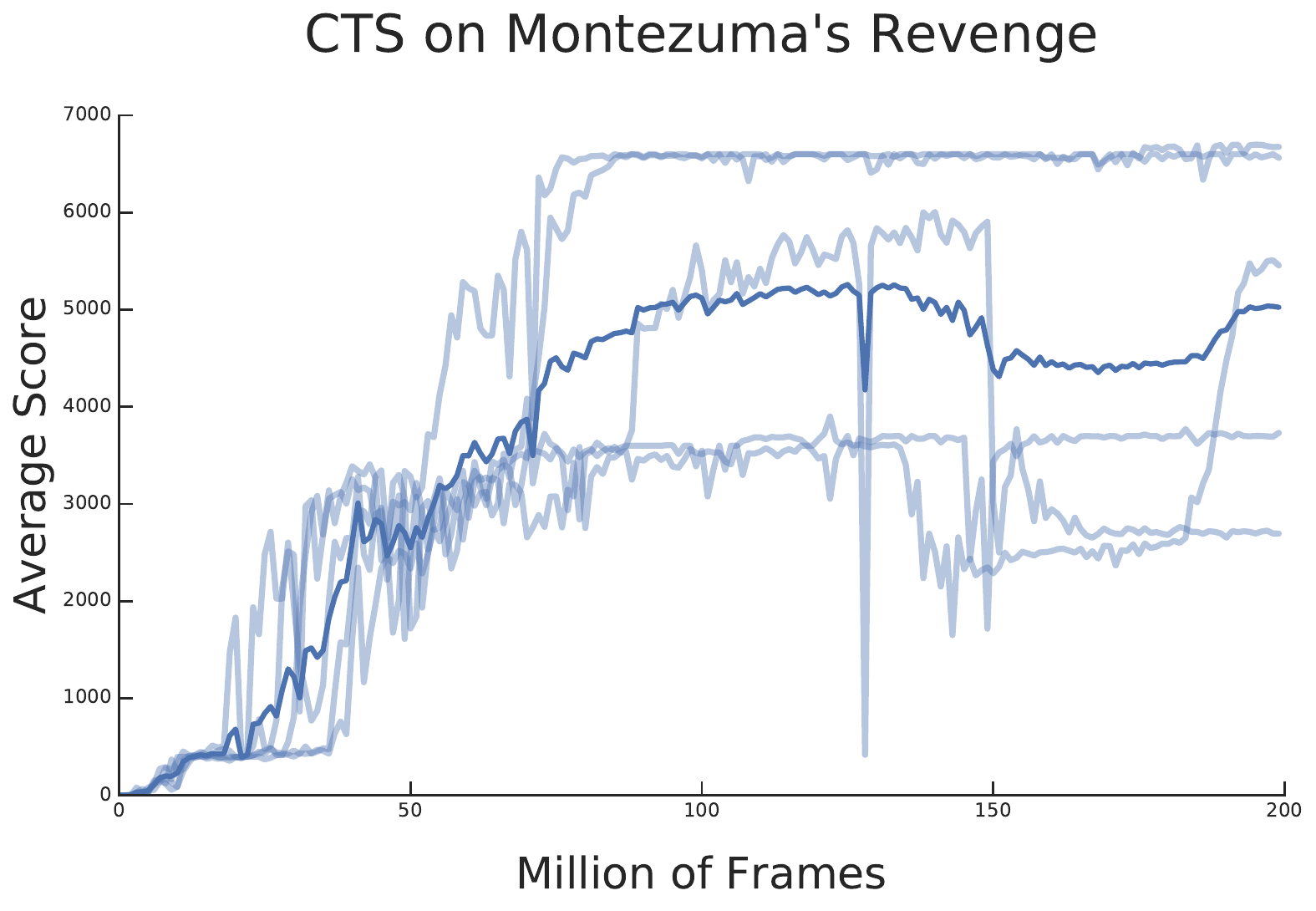}
  \label{fig:cts_montezuma}
\end{subfigure}%
\begin{subfigure}{0.33\textwidth}
    \centering
  \includegraphics[width=\linewidth]{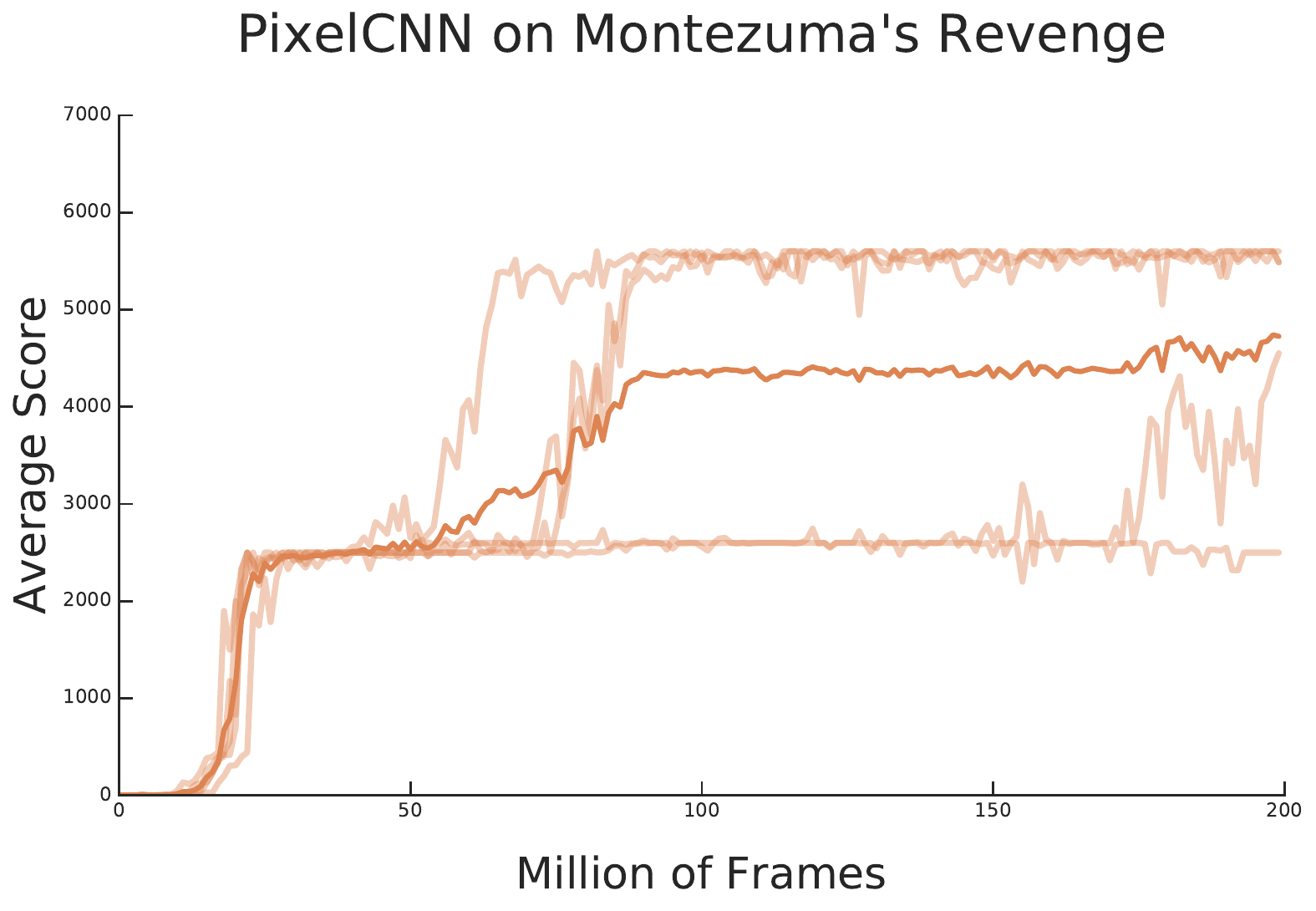}
  \label{fig:pixelcnn_montezuma}
\end{subfigure}%
\begin{subfigure}{0.33\textwidth}
\centering
  \includegraphics[width=\linewidth]{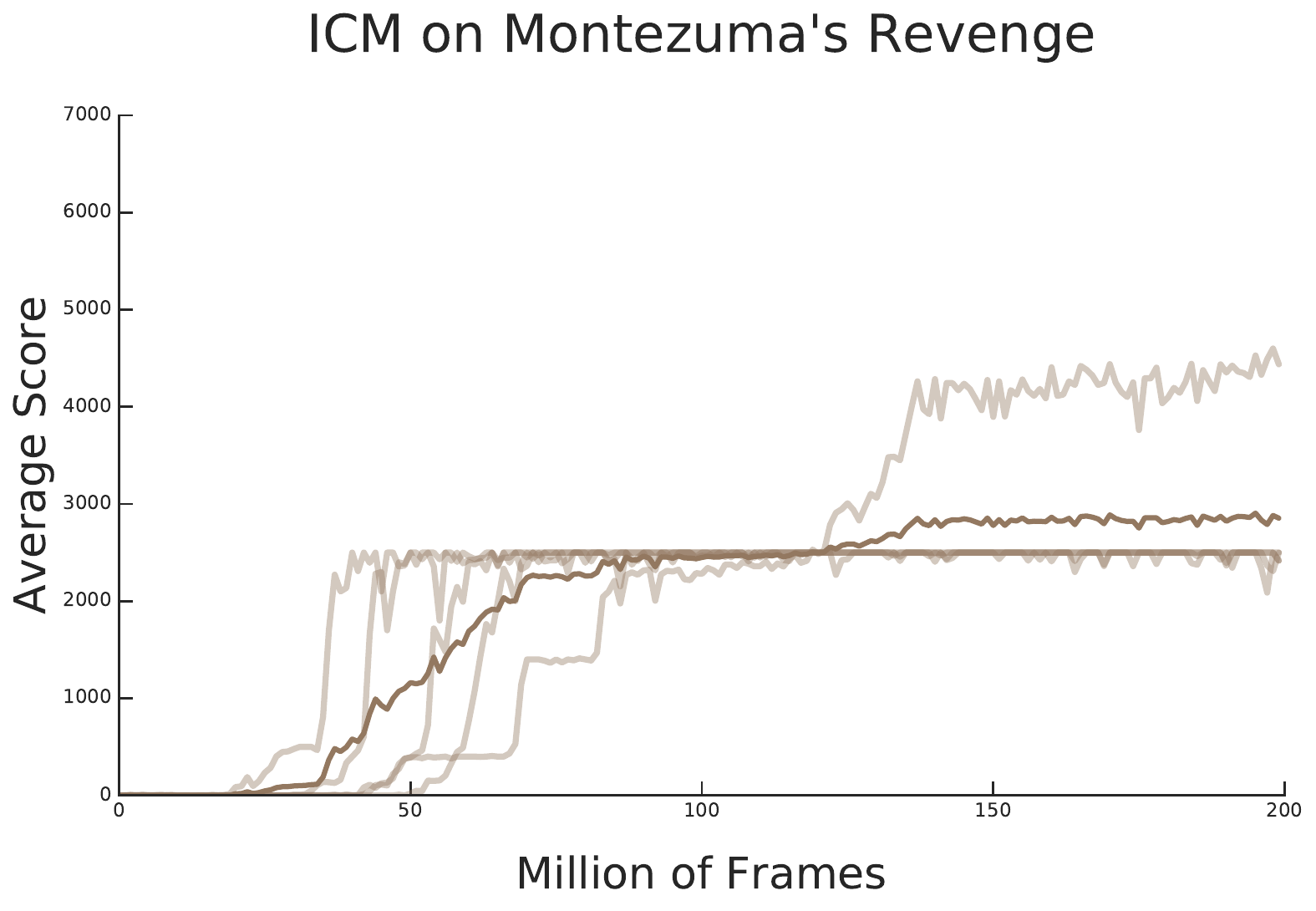}
  \label{fig:icm_montezuma}
\end{subfigure}%
\vspace{0.5mm}
\begin{subfigure}{0.33\textwidth}
\centering
  \includegraphics[width=\linewidth]{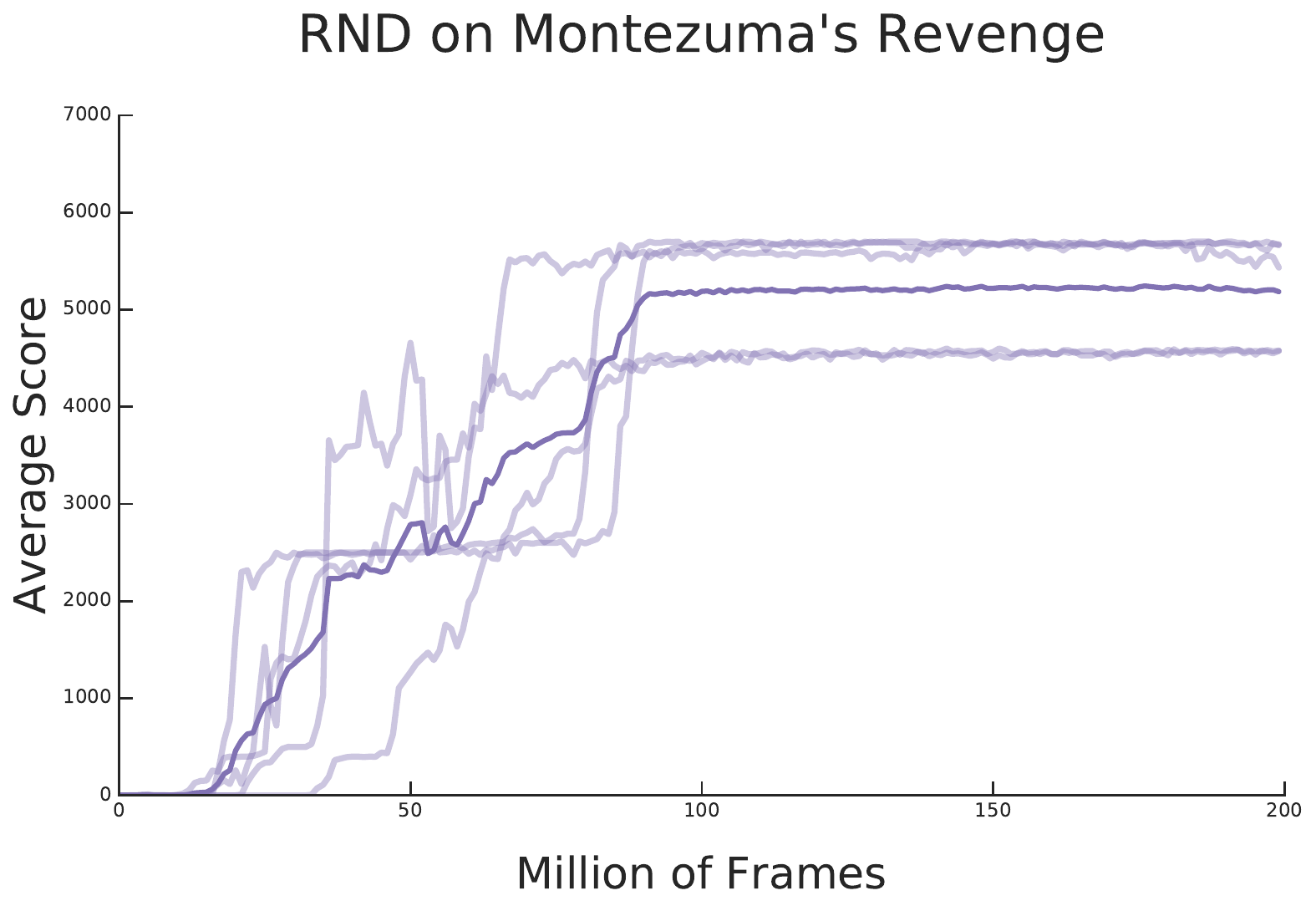}
  \label{fig:rnd_montezuma}
\end{subfigure}%
\begin{subfigure}{0.33\textwidth}
\centering
  \includegraphics[width=\linewidth]{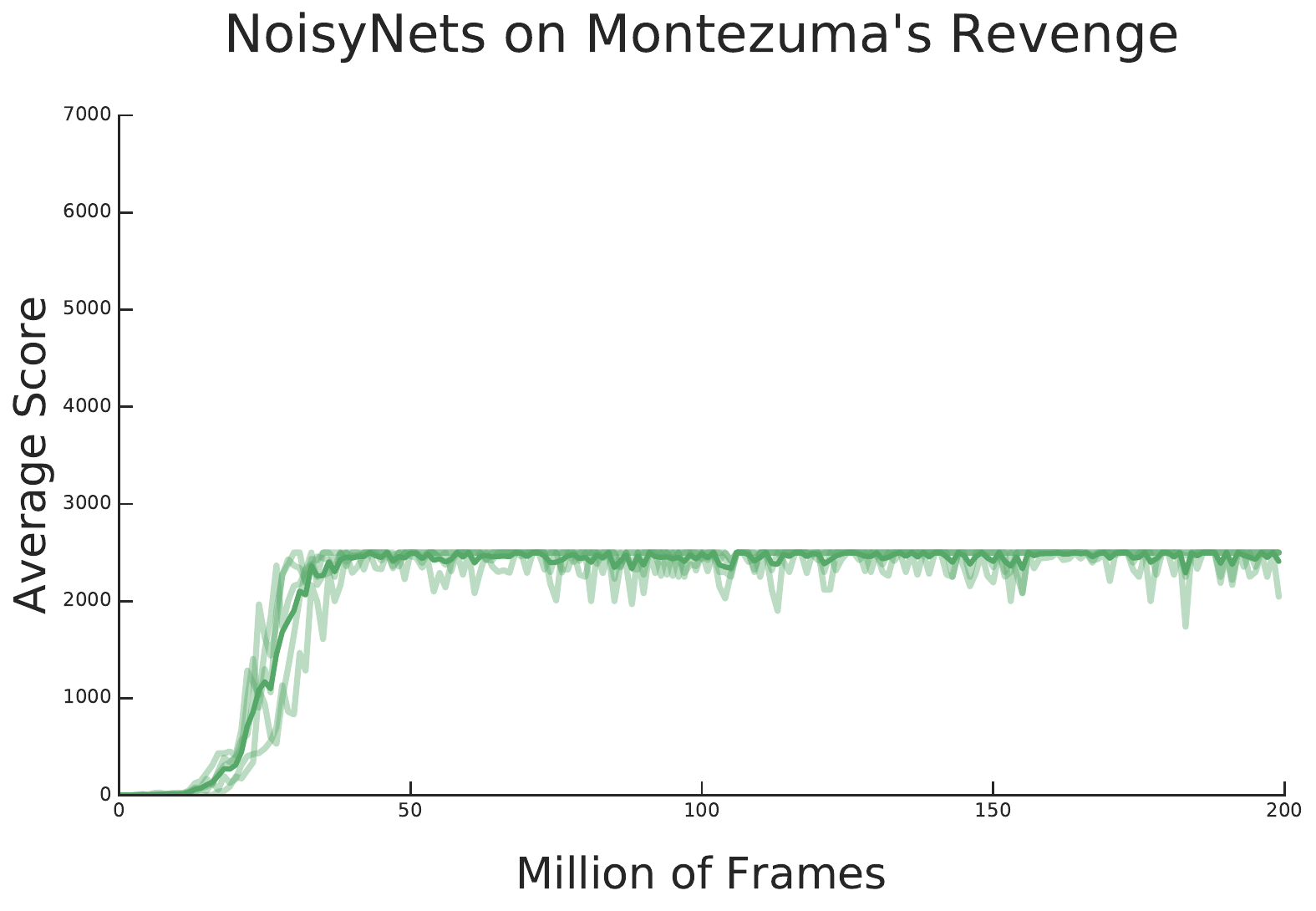}
  \label{fig:noisy_montezuma}
\end{subfigure}%
\begin{subfigure}{0.33\textwidth}
\centering
  \includegraphics[width=\linewidth]{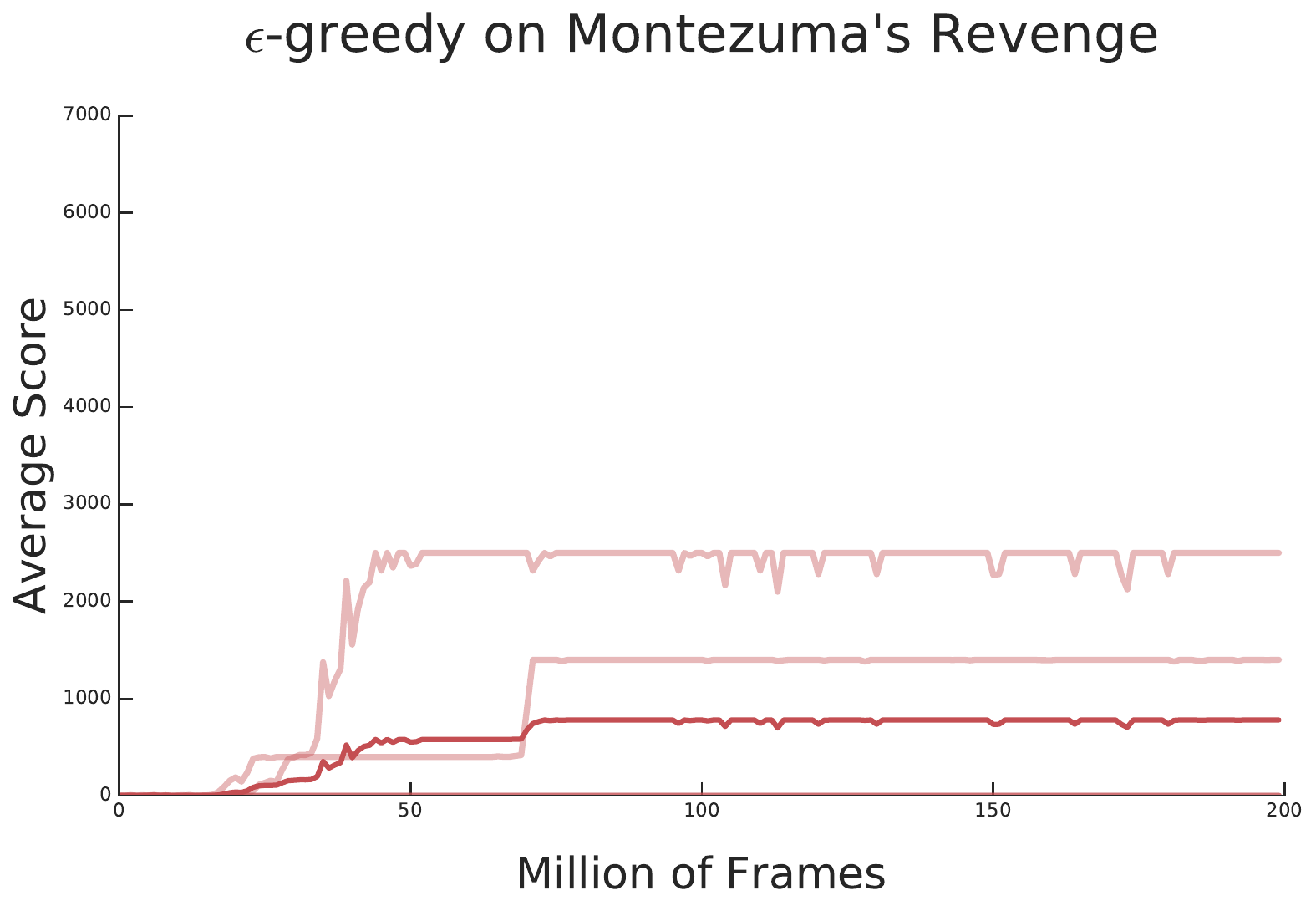}
  \label{fig:greedy_montezum}
\end{subfigure}%
\caption{Training curves on \montezuma}
\label{fig:training_curves}
\end{figure*}

The variance of the return on \montezuma{} is high because the reward is a step function, for clarity we also provide all the training curves in Figure~\ref{fig:training_curves}

\section{Hyperparameter tuning}
\label{app:hparam_tuning}

Except for NoisyNets, all other methods are tuned with respect to their final performance on \montezuma{} after training on 200 million frames on five runs.

\subsection{Rainbow and Atari preprocessing}

We used the standard architecture and Atari preprocessing from \citet{mnih2015human}. Following \citet{machado2018revisiting} recommendations we enable sticky actions and deactivated the termination on life loss heuristic. 
The remaining hyperparameters were chosen to match \citet{hessel2018rainbow} implementation.

\begin{center}
\begin{tabular}{ l|c} 
 \multicolumn{1}{c}{Hyperparameter} & Value \\
  \hline
 Discount factor $\gamma$ & 0.99 \\
 Min history to start learning & 80K frames \\ 
 Target network update period & 32K frames \\
 Adam learning rate & $6.25 \times 10^{-5}$  \\
 Adam $\epsilon$ & $1.5 \times 10^{-4}$ \\
 Multi-step returns $n$ & 3 \\
 Distributional atoms & 51 \\
 Distributional min/max values & [-10, 10] \\
\end{tabular}
\end{center}
Every method except NoisyNets is trained with $\epsilon$-greedy following the scheduled used in Rainbow with $\epsilon$ decaying from $1$ to $0.01$ over 1M framces.

\subsection{NoisyNets}
We kept the original hyperparameter $\sigma_0 = 0.5$ used in \citet{fortunato18noisy} and \citet{hessel2018rainbow}.

\subsection{Pseudo-counts}
We followed \citeauthor{bellemare2016unifying}'s preprocessing, inputs are $42 \times 42$ greyscale images, with pixel values quantized to 8 bins.

\subsubsection{CTS}
We tuned the scaling for $\beta \in \{ 0.5, 0.1, 0.05, 0.01, 0.005, 0.001, 0.0005, 0.0001 \}$ and found that $\beta = 0.0005$ worked best.

\subsubsection{PixelCNN}
We tuned the scaling factor and the prediction gain decay constant $c$. We ran a sweep with the following values: $\beta \in \{ 5.0, 1.0, 0.5, 0.1, 0.05 \}$, $c \in \{5.0, 1.0, 0.5, 0.1, 0.05 \}$ and found $\beta = 0.1$ and $c = 1.0$ to work best.

\subsection{ICM}
We tuned the scaling factor and the scalar $\alpha$ that weighs the inverse model loss against the forward model. We ran a sweep with $\alpha = \{0.4, 0.2, 0.1, 0.05, 0.01, 0.005 \}$ and $\beta = \{2.0, 1.0, 0.5, 0.1, 0.05, 0.01, 0.005, 0.001, 0.0005 \}$. We chose $\alpha = 0.005$ and $\beta = 0.005$ to work best.

\subsection{RND}
Following \citet{burda2018exploration} we did not clip the intrinsic reward while the extrinsic reward was clipped (we also found in our initial experiments that clipping the intrinsic reward led to worse performance). We tuned the reward scaling factor and Adam learning rate used by RND optimizer. We ran a sweep with $\beta = \{0.5, 0.1, 0.05, 0.01, 0.005, 0.001, 0.0005, 0.0001, 0.0005$ and $\text{lr} = \{ 0.001, 0.0005, 0.0002, 0.0001, 0.00005 \}$.
We found that $\beta = 0.0001$ and $\text{lr} = 0.0002$ worked best.
\end{appendices}
\end{document}